\documentclass[lettersize,journal]{IEEEtran}
\usepackage{amsmath,amsfonts}
\usepackage{algorithm}
\usepackage{array}
\usepackage[caption=false,font=normalsize,labelfont=sf,textfont=sf]{subfig}
\usepackage{textcomp}
\usepackage{stfloats}
\usepackage{url}
\usepackage{verbatim}
\usepackage{graphicx}
\usepackage{cite}
\usepackage{wrapfig}
\usepackage{bbm}
\usepackage{balance}
\usepackage{adjustbox}
\usepackage{multirow}
\usepackage{array}

\usepackage[utf8]{inputenc}
\usepackage{algpseudocode}
\usepackage{xcolor}
\newcommand{\coloredComment}[1]{\Statex \hspace*{5em} \textcolor{gray}{\% #1}}

\newcommand{\coloredCommentTwo}[1]{\Statex  \textcolor{gray}{\% #1}}
\newcommand{\coloredCommentThree}[1]{\Statex \hspace*{2em} \textcolor{gray}{\% #1}}
\begin{document}

\title{Minimally-intrusive Navigation in Dense Crowds with Integrated Macro and Micro-level Dynamics}
\author{
\IEEEauthorblockN{
Tong Zhou$^{1, \#}$,
Senmao Qi$^{2, \#}$, 
Guangdu Cen$^{2}$,  
Ziqi Zha$^{2}$,  \\
Erli Lyu$^{3}$, 
Jiaole Wang$^{2, *}$,
Max Q.-H. Meng$^{4, *}$, \IEEEmembership{Fellow, IEEE}
}

\thanks{$^{1}$ Tong Zhou is with Department of Electronic Engineering, The Chinese University of Hong Kong, Shatin, N.T., Hong Kong SAR, China.}
\thanks{$^{2}$ Jiaole Wang, Guangdu Cen, Ziqi Zha, Senmao Qi are with School of Mechanical Engineering and Automation, Harbin Institute of Technology (Shenzhen), Shenzhen, China, 518055.}
\thanks{$^{3}$ Erli Lyu is with Faculty of Applied Science, Macao Polytechnic University, Macao, China.}
\thanks{$^{4}$ Max Q.-H. Meng is with the Shenzhen Key Laboratory of Robotics Perception and Intelligence, and the Department of Electronic and Electrical Engineering, Southern University of Science and Technology, Shenzhen 518055, China, on leave from the Department of Electronic Engineering, The Chinese University of Hong Kong, Hong Kong, and also with the Shenzhen Research Institute of The Chinese University of Hong Kong, Shenzhen 518057, China.}
\thanks{$^*$ Corresponding authors: Jiaole Wang (e-mail: wangjiaole@hit.edu.cn),
Max Q.-H. Meng (e-mail: max.meng@ieee.org).}
\thanks{$^\#$ Tong Zhou and Senmao Qi contributed equally to this work.}
}


\markboth{Journal of \LaTeX\ Class Files,~Vol.~14, No.~8, August~2021}%
{Shell \MakeLowercase{\textit{et al.}}: A Sample Article Using IEEEtran.cls for IEEE Journals}

\maketitle

\begin{abstract}
In mobile robot navigation, despite advancements, the generation of optimal paths often disrupts pedestrian areas. To tackle this, we propose three key contributions to improve human-robot coexistence in shared spaces.
Firstly, we have established a comprehensive framework to understand disturbances at individual and flow levels. Our framework provides specialized computational strategies for in-depth studies of human-robot interactions from both micro and macro perspectives. By employing novel penalty terms—Flow Disturbance Penalty (FDP) and Individual Disturbance Penalty (IDP)—our framework facilitates a more nuanced assessment and analysis of the robot navigation's impact on pedestrians.
Secondly, we introduce an innovative sampling-based navigation system that adeptly integrates a suite of safety measures with the predictability of robotic movements. This system not only accounts for traditional factors such as trajectory length and travel time but also actively incorporates pedestrian awareness. Our navigation system aims to minimize disturbances and promote harmonious coexistence by considering safety protocols, trajectory clarity, and pedestrian engagement.
Lastly, we validate our algorithm's effectiveness and real-time performance through simulations and real-world tests, demonstrating its ability to navigate with minimal pedestrian disturbance in various environments.
% Lastly, the versatility and reliability of our algorithm have been validated through extensive simulations and practical testing. Demonstrating proficient navigation in both constrained geometries and open terrains, our approach strengthens the practical impact of autonomous robot navigation. The empirical results highlight the effectiveness of our method in reducing pedestrian disturbances and improving real-time navigation performance.This document describes the most common article elements and how to use the IEEEtran class with \LaTeX \ to produce files that are suitable for submission to the IEEE.  IEEEtran can produce conference, journal, and technical note (correspondence) papers with a suitable choice of class options. 
\end{abstract}


\section{Introduction}
\label{sec:introduction}

In the advent of smart urban ecosystems, the synergy between autonomous robots and human inhabitants is becoming increasingly prevalent \cite{robot_city_interaction, 7580764}. These robots, with a growing presence in our daily environments, are tasked with intricate operations such as navigating through hospital corridors to deliver supplies, thus maintaining the unobstructed movement of medical staff and patients. Concurrently, they enhance the efficiency of transit experiences by offering guidance, disseminating information, and handling luggage within the complex network of transportation nodes like airports and train stations \cite{6916032, 7578407, chen2019crowd}. Such integrations underscore the symbiotic roles robots are designed to play, operating in tandem with human activities.

However, the quest to enable robots to traverse densely populated spaces safely and effectively has largely been concentrated \cite{trautman2015robot, 8814288} on refining pathfinding algorithms and circumventing obstacles, with an acute focus on aligning robotic actions with social and cultural norms  \cite{mavrogiannis2019multi, poddar2023crowd, liu2021social, tai2018socially}. While these endeavors are pivotal for the evolution of robotics, they frequently fail to address the nuanced effects these autonomous agents have on the dynamics of human movement around them \cite{poddar2023crowd}. Significantly, there is a noticeable shortage of methodical assessments simultaneously concerning the macroscopic and microscopic perturbations induced by robotic agents—factors that are essential for the nuanced understanding and enhancement of robot navigational tactics.

In response to this gap, our research introduces a sophisticated, bifocal method designed to assess the implications of robotic agents on the flux of pedestrian traffic—spanning the broad-scale alterations to collective movement patterns and the fine-scale modifications to the trajectories of individual pedestrians. By adopting this comprehensive perspective, we can thoroughly evaluate the repercussions of robotic integration within densely populated settings. The validity and applicability of our approach are corroborated through a series of evaluations, encompassing both simulated environments and empirical experiments.
\begin{figure}[t]
    \centering
    \includegraphics[width=\columnwidth]{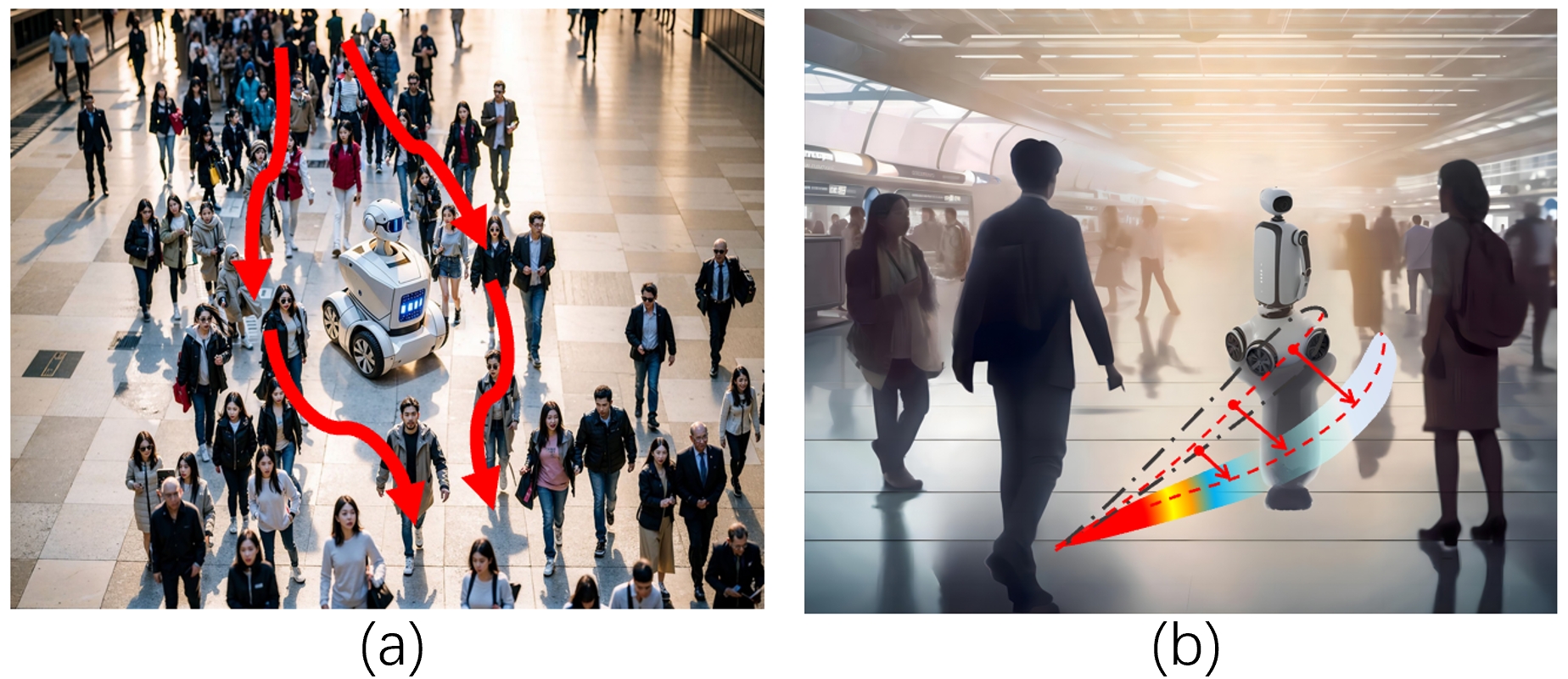}
    \caption{Disturbance in the navigation of robots and pedestrians encompasses two key aspects. (a) At a macroscopic level, the mere presence of a robot and its trajectory can exert influence on the flux of pedestrian flow, affecting its magnitude and direction. (b) At a microscopic level, the existence of a robot necessitates pedestrians to deviate from their desired paths, potentially resulting in increased travel distances. These issues not only contribute to congestion and safety concerns within the crowd but also give rise to the phenomenon known as the "freezing robot problem."}
    \label{fig:motivation}
\end{figure}

Our research advances three significant contributions:

\begin{itemize}

\item We establish a framework for understanding disturbances at two levels—individual and flow—providing specialized computational strategies for in-depth examinations of human-robot interactions from both micro and macro perspectives.
\item Our innovative sampling-based navigation system intricately weaves together a network of safety measures, enhances the predictability of robotic movements, and actively considers pedestrian awareness to refine the cohabitation of robots and humans in shared spaces.
\item We substantiate the versatility and reliability of our algorithm with both simulation and real-world tests, demonstrating proficient navigation in both constrained geometries and open terrains, thereby reinforcing the practical impact of our work on autonomous robotic navigation.
\end{itemize}

\section{Related Work}
\label{sec:related_work}
\subsection{Partial Motion Plannings}
In the evolving domain of Partial Motion Planning (PMP), safety and trajectory suboptimality in dynamic environments are of paramount concern. \cite{bouraine2012provably} introduces the Braking Inevitable Collision State (BICS) principle to enhance collision avoidance, which is foundational to passive safety. Trajectory planning methodologies, such as the RRT \cite{passive_safe_partial, provably_safe_motion}, have been developed to bolster safety during navigation amidst crowds. Nevertheless, these approaches may inadvertently constrain interaction dynamics with pedestrians and fall short on ensuring trajectory optimality.

Advancements in interactive navigation are evident in the work of \cite{sun2021move}, which utilizes a sampling-based approach to generate a spectrum of partial trajectories, thereby optimizing multi-agent navigation through a delicate balance of individual preferences and mutual accommodation to avert collisions. Similarly, \cite{mavrogiannis2019multi} devises an anticipatory algorithm that negotiates system path interactions by analyzing permutations of adjacent agents and forecasting braid words through temporal progression. This methodological foresight enables the selection of trajectories that are congruent with anticipated braid patterns, thereby refining navigational choices. However, these strategies presuppose the need for precise predictions of pedestrian behavior, a prerequisite that, if unmet due to unpredicted pedestrian noncompliance, could undermine algorithmic performance.

Addressing the complexities of trajectory decision-making, \cite{zhou2023towards} adopts Trajectory-level Monte Carlo Tree Search (MCTS), which scrutinizes partial trajectories at each node and amalgamates learning-based forecasts of their influences on surrounding agents with a pool of expert policy insights. Despite these sophisticated approaches, the challenge of bridging the simulation-to-reality divide persists, posing a persistent obstacle to the practical implementation of reinforcement learning-driven models.

\subsection{Crowd Reaction Modeling}
Crowd interaction modeling is a critical component in the field of robotics, particularly for the development of autonomous systems that operate within human environments. The ability of a robot to navigate through crowded spaces safely and efficiently hinges on its understanding of human behavior and crowd dynamics.
\subsubsection{Macroscopic Models and Microscopic Models}

Crowd representation can be divided into macroscopic representation, which considers crowd density, mass velocity, and energy as dependent variables of time and space, and microscopic representation, which treats each pedestrian as a particle with individually identified position and velocity. 

Macroscopic models prioritize the collective behavior of crowds. Dogbe's research \cite{dogbe2008numerical} examines the numerical solutions for pedestrian flow, focusing on a desired speed and avoidance of crowded areas. Maury and colleagues \cite{maury2010macroscopic} apply these principles to study crowd movements during evacuations, using gradient flow in the Wasserstein space to model how crowds interact with the environment and exits. These models are key for predicting overall pedestrian movement in scenarios where the crowd's flow is critical.

Microscopic models are of particular interest when it comes to robot-pedestrian interaction. They range from traditional models like the Social Force\cite{helbing1995social}, ORCA\cite{van2010optimal}, and their variations\cite{predhumeau2021agent}, to more recent learning-based methods\cite{gupta2018social, social_lstm, pang2021trajectory}. \cite{social_lstm}
is the first data-driven approach that replaces manual feature extraction with a "social pooling layer," enabling agents in close proximity to share state information and learn interactions between them. \cite{gupta2018social} incorporates GAN networks to encourage the generation of multi-modal trajectories. Subsequent work\cite{pang2021trajectory} aims to improve the accuracy of trajectory prediction tasks by considering additional factors and longer time horizons.

\subsubsection{Homogeneous Reaction Modeling and Heterogeneous Reaction Modeling}
we investigate whether pedestrians' reactions to robots are similar to their reactions to other pedestrians, termed as homogeneous, or if they are treated separately, modeling pedestrians' trajectory changes individually, referred to as heterogeneous. 

Homogeneous reaction models play a pivotal role in multi-agent navigation by assuming that pedestrians react to robots in the same way they would to other humans \cite{mavrogiannis2019multi, sun2021move, trautman2015robot}. These models are crucial for enabling robots to navigate densely populated areas in a manner that is harmonious with human pedestrian behavior.
In the work by Mavrogiannis et al. \cite{mavrogiannis2019multi}, an algorithm predicts the final locations of agents based on their goals, guiding robots to select paths that are congruent with these predicted positions. Trautman's Interaction Gaussian Process model \cite{trautman2015robot}, informed by pedestrian data, integrates the robot into the crowd by mirroring pedestrian interaction patterns.
Sun et al. \cite{sun2021move} extend this idea by treating the interactions between robots and pedestrians on an equal footing. Their algorithm employs an iterative best response technique, constructing a reflective model of pedestrian behavior in the decision-making process. This model strives to minimize the differences between actual and preferred movement patterns while simultaneously reducing the likelihood of collisions, fostering a spirit of cooperation. The approach's validity was tested against the ETH and UCY pedestrian dataset, which confirmed that the interactions occurred as the model predicted.
By adopting the premise that pedestrians respond to robots as they would to other people, these models enable robots to seamlessly integrate into human crowds, maintaining natural and predictable interactions. 

Salzmann et al. \cite{salzmann2021trajectron} and Predhumeau et al. \cite{predhumeau2021agent} recognize that the dynamics between autonomous vehicles and pedestrians are inherently different from those between pedestrians due to variations in size and kinematics. Predhumeau et al. \cite{predhumeau2021agent} adjusted the social force model to better represent the interactions that occur between pedestrians, both as individuals and in groups, when they encounter vehicles.
In the approach proposed by Salzmann et al. \cite{salzmann2021trajectron}, distinct labels are assigned to pedestrians and vehicles during the network training phase, with the aim of accurately predicting the unique ways each entity interacts with the others. This distinction is critical for identifying and modeling the different interaction patterns that occur between pedestrians and vehicles, ensuring that the behavior of each can be anticipated and understood in the context of their coexistence.

\subsection{Disturbance Modeling}
Research extending beyond how pedestrians perceive robots includes defining metrics for the impact robots exert on humans. The "winding number," introduced by Mavrogiannis et al. \cite{mavrogiannis2022winding} and Roh et al. \cite{roh2020multimodal}, quantifies the interaction topology during passing maneuvers, with positive values indicating a natural counterclockwise movement that reflects human passing behavior.
The concept of social momentum, as explored in \cite{social_momentum}, serves a similar purpose where the sign of momentum infers alignment with expected trajectory patterns. Both methods, however, only address topological aspects without delving into detailed measures of influence.

Tolstaya et al. \cite{tolstaya2021identifying} employ KL-divergence to compare the change of trajectory distribution of an agent under the influence of the other against its usual behavior. This approach is limited to pairwise interactions and isn't suited for complex engagements involving multiple pedestrians and robots.
In research by Fan et al. \cite{fan2021crowddriven}, metrics for crowd interaction include costs for 'lubricating' (easing movement through the crowd) and 'resistance' (impediments faced). These gauge the robot's effect on a crowd's flow but overlook the temporal aspect, which is critical in dynamic environments.
Bresson et al.\cite{bresson2019socially} propose the 'cluster dodge distance' to measure the clearance a robot maintains from groups, aiding in devising crowd-bypassing strategies. Like the previous methods, this metric also lacks a temporal component, challenging its efficacy in time-sensitive contexts.

%%%%%%%%%%%%%%%%%%%%%%%%%%%%%%%%%%
\section{PROBLEM FORMULATION}
\begin{figure*}[t]
    \centering
    \includegraphics[width=\textwidth]{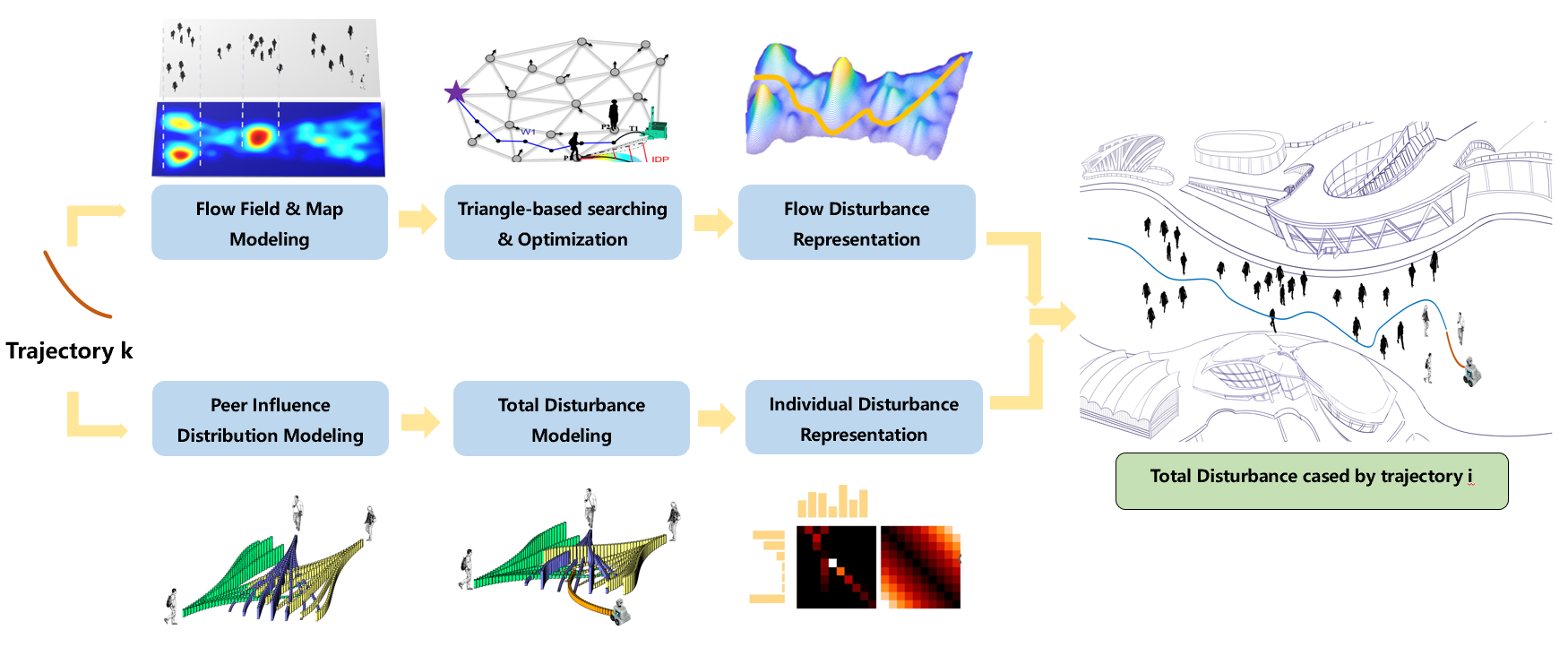}
    \caption{Disturbance modeling aims to analyze the impact of a given robot trajectory on pedestrian flow, considering both macroscopic and microscopic aspects. Macroscopically, the trajectory's endpoint is evaluated for its ability to facilitate smooth passage through the crowd while minimizing flux variations. This can be achieved through techniques such as flow map modeling, triangle-based search, and optimization. On the microscopic level, the influence of the trajectory on the distribution of surrounding pedestrian trajectories is assessed. By simultaneously modeling scenarios with and without the presence of the robot executing the specified trajectory, the push distance metrics are computed. Ultimately, these components of disturbance are integrated to provide a comprehensive assessment of the trajectory's overall impact on the pedestrian flow.}
    \label{fig:disturbance_modeling}
\end{figure*}
\label{sec:problem_formulation}
We first define two levels of disturbances caused by the robot to the pedestrian flow.
From the macroscopic perspective, we quantify the disturbance as the overall impact on pedestrian flow caused by the robot's presence. This refers to the extent of the impact, in magnitude and direction, that the robot's presence has on pedestrian flow compared to when the robot is absent.
On the microscopic level, the disturbance is defined as the discrepancy between the future trajectory distributions of individual pedestrians with and without the robot. Specifically, this quantifies how much additional displacement a pedestrian must make to avoid the robot, compared to the scenario where no robot is present.
To articulate this problem and our objectives more precisely, we first define some basic tuples and symbols.
The problem of minimal intrusive navigation in crowded environments is defined by a tuple $\mathcal{<X, A,  H, T, P, M, J>} $, where 
\begin{itemize}
  \item $\mathcal{X} \in \mathbb{R}^2$ denotes the workspace that contains the robot and pedestrians. Without loss of generality, we assume that there are a total of n+1 agents in the environment, including one robot and n pedestrians. We index them as $\{R, 1, 2, ..., n\}$.
  \item $\mathcal{A} \subset {X}$ is a closed region occupied by the robot in the workspace. $\mathcal{A}(t)$ denotes the mapping from the time space to the workspace.
      The motion model is defined by a state function of the robotic system,
    \begin{equation}
    	\dot{\mathbf{x}}=f(\mathbf{x}, \mathbf{u}),
    \end{equation}
    with invariant constraints 
    \begin{equation}
    	h(\mathbf{x}, \mathbf{u})=0,\; h(\mathbf{x}, \mathbf{u}) \leq 0,
    	\label{motion-c}
    \end{equation}
    where $\mathbf{x}$ denotes the robot state, $\mathbf{u}$ denotes the control input, and Eq. \eqref{motion-c} defines a set of motion constraints that are invariant to the environment.
  \item $\mathcal{H}$ denotes a set of pedestrians $\mathcal{H}_i, i \in \{1, ,2, ..., n \}$ moving in the workspace. Collision happens if the robot $\mathcal{A}$ intersects $\mathcal{H}_i$, i.e., $\mathcal{A}(t) \cap \mathcal{H}_i(t) \neq \emptyset$.
  \item $\mathcal{T}$ denotes the trajectory space of an agent. $\{ \tau_i \}_J=\{ \tau_{i, 1}, \tau_{i, 2}, ..., \tau_{i, J} \}$ represents a trajectory bundle composed of J trajectories for agent $i$, where the $j$-th trajectory $\tau_{i,j} \in \mathbb{R}^{2 \times T}, j= \{1,2,.., J\}$, indicating that it includes the state information of the agent $i$ for T time steps.
 \item $\mathcal{P}$ is the distribution space, where each element $p(\tau): \mathcal{T} \rightarrow \mathbb{R}_0^+$ is a probability density function that maps trajectory space $ \mathcal{T}$ to the non-negative real domain, and each element satifies.
 \begin{equation}
 \int_{\mathcal{T}}p(\tau)d \tau = 1
 \end{equation}
Strictly speaking, we can have three different types of distributions, namely:
\begin{enumerate}
\item No-Interference Distribution (NID): This distribution represents the preference distribution when pedestrians are not subjected to any interference, whether from other pedestrians or robots.
\item Peer-Influence Distribution (PID): This distribution represents the preference distribution when pedestrians mutually influence each other, but without any interference from robots.
\item Total-Influence Distribution (TID): This distribution includes the preference distribution when pedestrians are influenced both by other pedestrians and the presence of robots. The distribution exhibits sensitivity to the divergent future decisions of the robot, thereby resulting in consequential variations in its behavior.
\end{enumerate}
\item Map representation $\mathcal{M}$ defines a model for the environment, which separates the free space $\mathcal{M}_{free}$ from static obstacles $\mathcal{M}_{obstacle} $. In addition to static maps, we also derive spatio-temporal maps of pedestrian flow, denoted as $\mathcal{M}_{flow} \subset \mathcal{M}_{free}$, which represent the flux of pedestrian flow at specific locations and times.

\item $\mathcal{J}$ is the objective function. Assuming that the vehicle start at $t_s$ and reach the goal state at $t_g = t_s + T$. The boundary value specifies the start state $x(t_s) = x_s$ and goal state $x(t_g) = x_g$ of a planning process.

The objective function includes a terminal cost L and a running cost C,
\begin{equation}
    J = L(x(t_g)) + \int_{t_s}^{t_g}{C(x(t), u(t))dt}
\end{equation}

We represent the entire objective function in the form of partial motion planning, where the first part describes the macroscopic impact of the local destination $x(t_g)$ on pedestrian flow in the future. The latter part describes the influence of the robot on individual pedestrians within T time steps when executing $[u(t_s), u(t_s+1), ..., u(t_g)]$.

\end{itemize}

\section{Method Overall}
\label{sec:method_overall}
\subsection{Intention-Topography Triangulation}
\begin{figure}[t]
    \centering
    \includegraphics[width=\columnwidth]{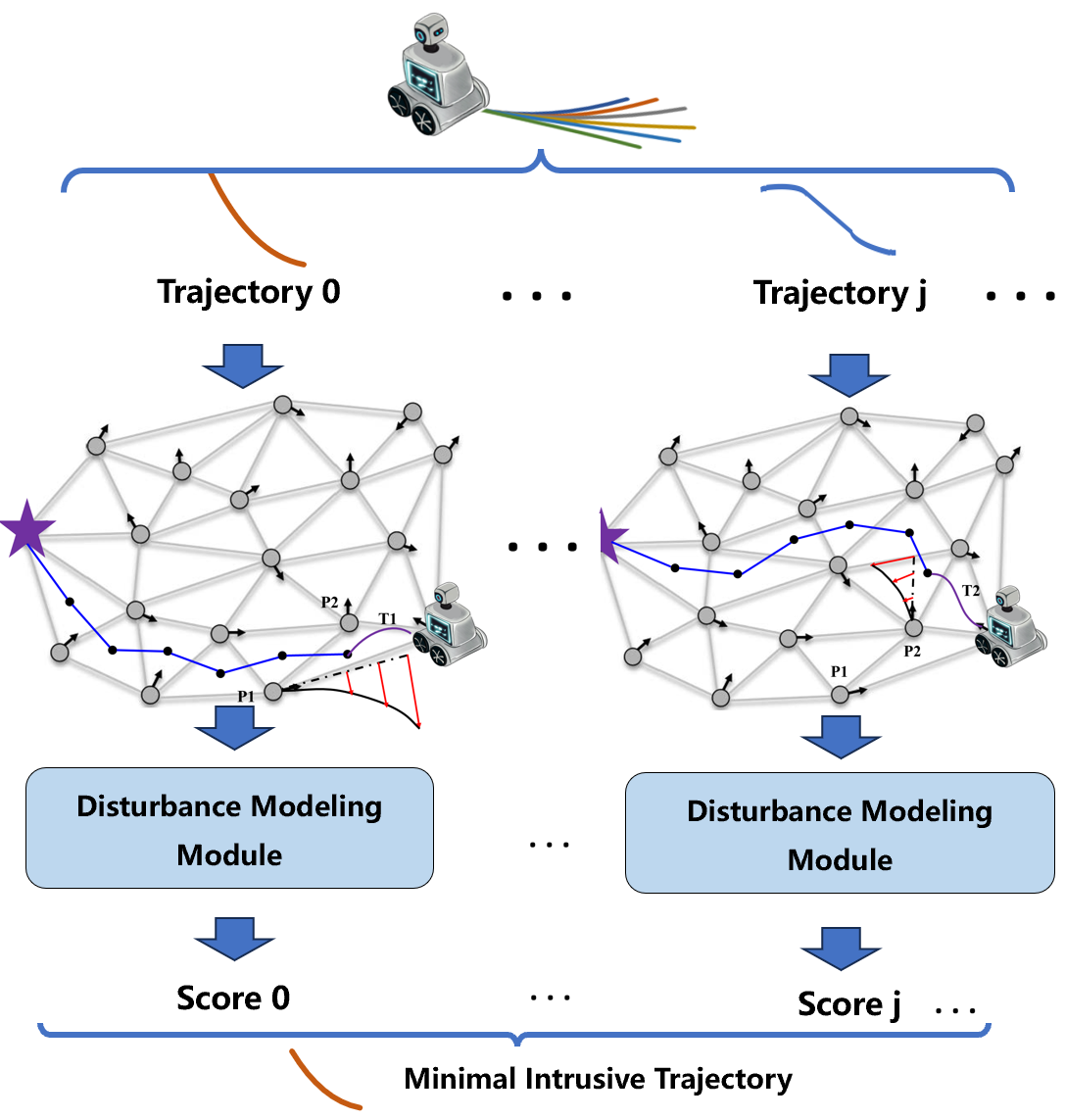}
    \caption{Intention-Topography Triangulation and Sampling-based Trajectory Generation Phase. It aims to efficiently generate trajectories by partitioning the triangle into phases that accommodate diverse intentions and topological structures. Multiple trajectories with fixed horizons are sampled for each phase, and disturbance calculations are performed to evaluate their impact. The trajectory with the lowest disturbance is selected as the optimal choice.}
    \label{fig:sampling-based-method}
\end{figure}
In navigating dense crowds, we present our innovative Intention-Topography Triangulation (ITT) approach. This method perceives pedestrians as nodes within a triangulation network, where the triangle faces act as 'Insertion Areas' signifying different navigation intents, and the nodes capture the topographical nuances of human movement.

Thus, ITT exploits the dual benefits of the triangulation network. It furnishes the robot with clear 'insertion' zones for navigation and a tangible understanding of moving between two pedestrians (crossing two triangle edges). Furthermore, it paves the way for future IDP modeling with search functions based on triangular nodes. Overall, ITT proves to be an effective strategy for navigating intricate, densely populated environments.

\subsection{Sampling-based Trajectory Generation and Evaluation}
After obtaining the triangle, we generate trajectories based on motion primitives within each face region of the triangle. Here, the duration of each trajectory is fixed, and its endpoint corresponds to the face of the triangle.

For each sampled trajectory, we model both macro-level and micro-level disturbances. On a micro-level, we measure the extent to which executing this trajectory would cause deviations in the walking paths of surrounding pedestrians, as introduced primarily in Section \ref{sec:individual_disturbance_modeling}. On a macro-level, we assess the degree of interaction with the macro-level pedestrian flow that needs to be traversed from the trajectory's endpoint to the final goal, as described mainly in Section \ref{sec:flow_disturbance_modeling}.

Once we have determined the overall impact of each trajectory on the pedestrian flow, we use it as an additional cost to score each trajectory. The trajectory with the highest score is selected as the final trajectory for execution by the robot.

\section{Individual Disturbance Modeling}
\label{sec:individual_disturbance_modeling}

In the complex environment of autonomous navigation, understanding and predicting the behavior of pedestrians is crucial. 
The robot's disturbance within a crowd encapsulates a complexity surpassing that of a 2-player game. 
The approach in work \cite{tolstaya2021identifying}, which models trajectory distributions merely as being either influenced by the robot or unaffected by anybody, is insufficient.
It overlooks the subtle yet significant interactions among pedestrians that shape trajectory distributions, even without robotic interference. To accurately capture pedestrian dynamics, these interactions must be incorporated into the model. Additionally, when considering robotic influence, it is critical to include both the direct effects, such as pedestrians adjusting their paths to avoid the robot, and the indirect effects, where one pedestrian's avoidance maneuver (e.g., pedestrian A) triggers trajectory changes in others (e.g., pedestrian B).
% It neglects the intricate pedestrian-to-pedestrian interactions, even in the robot's absence. When the robot's presence is factored in, it is imperative to consider both its direct impacts (pedestrians actively avoiding the robot) and its indirect effects (pedestrian A evading the robot, thereby influencing pedestrian B's movement).
% Pedestrians do not move in isolation; their trajectories are influenced by the presence and movement of others, including autonomous robots. 
In this work, we delve into the modeling of pedestrian behavior under varying degrees of influence from other pedestrians and robots.

\begin{figure}[hbpt]
    \centering
    \includegraphics[width=\columnwidth]{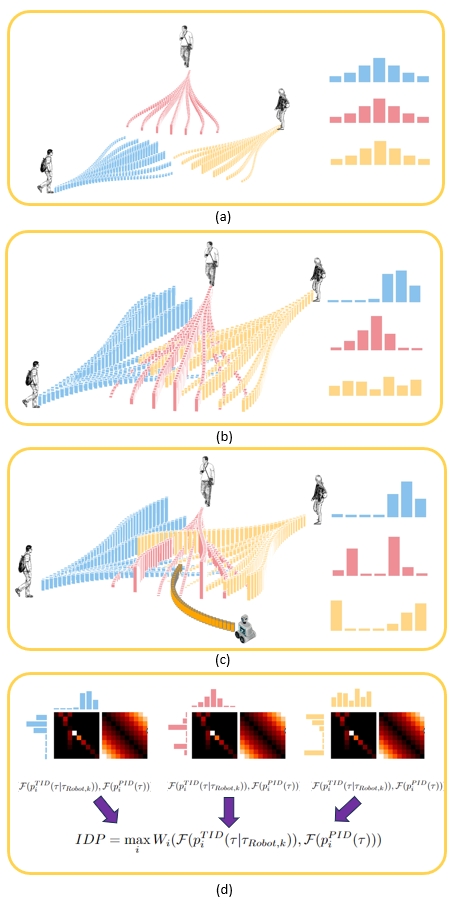}
    \caption{The figure delineates three distinct pedestrian distribution models. (a) No-Interference Distribution: Pedestrians navigate solely based on personal intentions, ignoring the potential interactions with other agents, resulting in trajectories that are optimally direct and linear. (b) Peer-Interference Distribution: This model reflects pedestrian trajectories where only interactions with other pedestrians are considered, effectively omitting robotic entities. According to multi-agent game theoretic principles, pedestrians adjust their paths to accommodate and avoid collisions with peers, while also navigating environmental constraints like physical terrain features. (c) Total-Interference Distribution: In this comprehensive model, pedestrians account for both other pedestrians and robots when selecting their paths. Trajectories become dynamic obstacles in the spatio-temporal domain, influencing the continual adjustment of pedestrian distributions. (d) The 'Wasserstein Distance' metric is used to quantify changes in a pedestrian's distribution, with the term 'Individual Disturbance' denoting the most pronounced distribution shift resulting from external influences.}
    \label{fig:idp}
\end{figure}

We first explore the theoretical underpinnings of multi-agent interactions, specifically through the lens of Game Theory. This provides the groundwork for understanding the interaction dynamics among multiple agents.
Subsequently, we focus on how pedestrians' trajectories evolve under different conditions. We start with a 'No Interference' scenario, capturing the initial intent of pedestrians when there are no external influences. We then consider the 'Peer-Influence Distribution', where pedestrians adjust their trajectories in response to other pedestrians but disregard the presence of the robot. Finally, we examine the 'Total-Influence Distribution', where pedestrians adapt their trajectories considering both the movements of other pedestrians and the robot.
The entire process can be visualized in Figure
\ref{fig:idp}.

The following sections will delve into each of these areas, shedding light on the complex interplay of influences that shape pedestrian behavior in an autonomous navigation context.

\subsection{Multi-agent Game Theory}
In the realm of multi-agent interactions, Game Theory provides a solid foundation for modeling and understanding the dynamics at play. In this section, we delve into the Multi-agent Game Theory, illustrating how it serves as a theoretical basis for pedestrian interference modeling. This foundation will pave the way for the subsequent sections, where we will explore different trajectory distributions under various conditions.
Each agent aims to minimize potential collisions while maintaining fidelity to its original intent, creating a multi-objective optimization problem \cite{sun2021move}. 
To navigate this complex interaction landscape and minimize potential conflicts, we need to define a set of penalties or costs associated with possible collisions. This leads us to the concept of 'Collision Penalty Representation'.
\subsubsection{Collision Penalty Representation}
The 'Collision Penalty Representation' provides a method to quantify the potential expenses resulting from collisions in our multi-agent game model. Here, we define three specific penalties:
\begin{itemize}
    \item \textbf{Coupled Trajectory Overlap Expense}:
   This penalty is derived from any two sampled trajectories for any two agents. It quantifies the degree of overlap between two trajectories in terms of space and time. We use $\psi$ to denote this cost. We use $\psi$ as the notation for it.
    \begin{equation}
    \psi(\tau^{(i)},\tau^{(j)}) : \mathcal{T} \times \mathcal{T} \rightarrow \mathbb{R}^+  
    \end{equation}
    \item \textbf{Expected Collision Penalty}: 
This penalty signifies the aggregate of all possible trajectories based on the distribution in the trajectory space. In essence, it represents the expected value of the collision penalty function over all feasible pairs of trajectories. The weight of each pair of trajectories is determined by the distribution in the trajectory space.
\begin{equation}
    c(p_i, p_j) = \int_{\mathcal{T}} \int_{\mathcal{T}} \psi(\tau^{(i)},\tau^{(j)}) p_i(\tau^{(i)})p_j(\tau^{(j)})d{\tau^{(i)}}d{\tau^{(j)}}
\end{equation}
\item \textbf{Joint expected collision penalty}:
For $n$ pedestrians, the aggregate anticipated collision penalty is defined as follows:
\begin{equation}
\label{eq:joint_expected_collision_penalty}
    J_c(p_1, p_2, ..., p_n) = \sum_{i=1}^n\sum_{j=1}^{n, j\neq i}c(p_i,p_j)
\end{equation}
\end{itemize}
% Compared to approaches that treat robots as homogeneous agents and include the probability distribution of robot trajectories in collision penalties \cite{sun2021move}, our method only accounts for the probabilistic distribution of pedestrian trajectories in penalties, treating robot paths as dynamic obstacles to be avoided by pedestrians. This simplifies the interaction model, reducing the infinite regress issue in reciprocal planning \cite{che2020efficient}.

In our work, we model the interactions within a multi-agent system primarily referencing the approach in \cite{sun2021move}, which encompasses modeling interactions among pedestrians as well as between pedestrians and robots. However, diverging from \cite{sun2021move}, we do not treat the robot as a member of the multi-agent intelligence, nor do we optimize its probabilistic distribution in tandem with the pedestrians. Instead, we consider a specific trajectory of the robot as a dynamic obstacle that the pedestrians must avoid. We solely evaluate the cost incurred by pedestrians to circumvent the robot, assuming that the robot steadfastly follows a predetermined trajectory without considering intermediate evasive maneuvers. This simplifies the interaction model, reducing the infinite regress issue in reciprocal planning \cite{che2020efficient}.

\subsubsection{Iterative Best Response}
The Iterative Best Response (IBR) is a technique we utilize to address optimization challenges among multiple participants \cite{9112709, 9329208}. This is predicated on the optimal response strategy in game theory. 
Each participant fine-tunes their outcome within a single iteration, under the assumption that the strategies of the other participants are known and immutable.
A new iteration cycle commences once every participant has had optimized his strategy. This iterative process of optimization enables participants to continually modify their strategies, progressively converging to an optimal solution.

\paragraph{Sequential Iterative Variational Update}
Given the cost function, the optimization process for agent $i$ in the $k$-th iteration is as follows:
\begin{equation}
\label{eq:sivu}
    p_i^{k+1}(\tau) = \arg min_p \{ D_{KL}(p||p_i^{(k)} + \overline{c_i}^{(k)}(p) \}
\end{equation}
where
\begin{equation}
\begin{split}
         \overline{c_i}^{(k)}(p) &= \sum_{j=1}^{i-1}c(p, p_j^{(k+1)}) + \sum_{j=i+1}^{n}c(p, p_j^{(k)}) \\
     &= \int_{\mathcal{T}}p(\tau)\overline{\gamma_i}^{(k)}(\tau)d\tau
\end{split}
\end{equation}

\begin{equation}
\begin{split}
    \overline{\gamma_i}^{(k)}(\tau) &= \sum_{j=1}^{i-1}\int_{\mathcal{T}}\psi(\tau, \tau^{(j)})p_j^{(k+1)}(\tau^{(j)})d\tau^{(j)} \\
    &+ \sum_{j=i+1}^{n}\int_{\mathcal{T}}\psi(\tau, \tau^{(j)})p_j^{(k)}(\tau^{(j)})d\tau^{(j)}
\end{split}
\end{equation}

The overall formula contains two components. The first element is the KL-divergence between preferences across two iterations, ensuring limited deviation from the initial intentions of the pedestrians. The second component is the estimated joint expected collision penalty, optimized to facilitate cooperation by conceding preferred spaces.

\paragraph{Solution to Iterative Best Response}
Referring to the research articulated in \cite{sun2021move}, it is shown that the optimal solution for Equation 1 is expressed by:

\begin{equation}
    p_i^{(k+1)}(\tau) = \frac{p_i^{(k)}(\tau) \cdot \exp{\overline{\gamma_i}^{(k)}(\tau)}}{\int_{\mathcal{T}}p_i^{(k)}(\tau) \cdot \exp{\overline{\gamma_i}^{(k)}(\tau)}d\tau}
\end{equation}
This established finding provides a pivotal reference point for our further discussions and analyses.
In addition, to address the 'curse of dimensionality' encountered when calculating the integral term in the preference update, a common practice is to use sampling-based methods. These techniques leverage Monte Carlo simulations to estimate the numerical value of the integral.

\subsection{No-Interference Distribution}
Transitioning from the theoretical underpinnings, we delve into a 'No Interference Distribution' scenario, devoid of external interventions. Here, we aim to map out the original preference distribution of pedestrians by sampling a set of trajectories and assigning weights to them, thereby establishing a baseline understanding of pedestrian behavior without considering their willingness to deviate from preferred trajectories to accommodate other agents.

For every pedestrian $i$, their historical trajectory, denoted as $[x_{-T}^{(i)}, ..., x_{-1}^{(i)}]$, can be observed. This allows us to generate predictions for future trajectories, represented by $p_i^{(NID)}(\tau) = p(\tau^{(i)}=\tau|x_{-T:-1}^{(i)}) \in \mathcal{P}$.
In the context of this research, we employ a Gaussian process model to encapsulate the agents' preferences, expressed as $p_i^{(NID)}(\tau) = \mathcal{N}(\tau|\mu_i, \Sigma_i)$. Here, $\mu_i$ signifies the primary direction of the pedestrian's desired intent, while $\Sigma_i$ delineates the dispersion of the distribution.

Adhering to the approach delineated in \cite{sun2021move}, we likewise generate $m$ trajectory samples from each agent $i$'s original distribution, yielding their initial weights. 
\begin{equation}
    \{\boldsymbol{\tau}_{i} \}^{(NID)} \sim p_i^{(NID)}(\tau)
\end{equation}

\begin{equation}
    \{\boldsymbol{\tau}_{i}\}^{(NID)} = \{  \boldsymbol{\tau}_{i,1}^{(NID)}, \boldsymbol{\tau}_{i,2}^{(NID)}, ..., \boldsymbol{\tau}_{i,m}^{(NID)} \}
\end{equation}

where each sample $\boldsymbol{\tau}_{i,j}^{(NID)}$ indicates the $j$-th sample of agent $i$, and it consists of two components: trajectories and their corresponding weights.

\begin{equation}
    \boldsymbol{\tau}_{i,j}^{(NID)} = (\tau_{i,j}, w_{(i,j)}^{(NID)})
\end{equation}

The weights signify the preference level of pedestrians for a specific trajectory. For the no-interference distribution, we set each weight to 1 as $w_{(i,j)}^{(NID)}=1$. In later calculations of the two distributions, we will maintain the same trajectories as in the no-interference distribution, while adjusting their weights.

\subsection{Peer-Influence Distribution}
Building on the 'No Interference Distribution', we establish the 'Peer-Influence Distribution' that takes into account the influence of other pedestrians on individual intentions and their propensity to shift their preferred trajectories for cooperative interactions, while intentionally excluding the robot's influence.

This approach is critical to maintaining minimal intrusion. By creating a unique 'parallel spacetime', we simulate and examine pedestrian interactions in the absence of a robot. This simulation provides a baseline estimation of pedestrians' natural interactions and gamesmanship, serving as a foundation for understanding and predicting how the introduction of a robot might alter these dynamics. Thus, it allows us to design robotic behaviors that harmonize with, rather than disrupt, the natural flow of pedestrian movement.

\subsubsection{Penalty Representation}

Within the framework of multi-agent game theory, a target pedestrian's interaction with others hinges on the inclusion of these agents' trajectories in their cost function. In our 'Peer-Influence Distribution' model, we deliberately exclude the robot from this cost function, focusing solely on non-robot entities. This approach allows us to analyze pedestrian dynamics in a robot-absent scenario.

\begin{equation}
\begin{split}
      \psi(\tau^{(i)},\tau^{(j)}) =& c_{ped} \max_{t} \gamma^t \cdot e^{\{-b \cdot [d(\tau^{(i)}(t), \tau^{(j)}(t)) - th_{peer}]  \}} \\
      &+ c_{obs} \cdot \mathbbm{1}_{\{{\tau^{(i)} \cap M_{obs} \neq \emptyset}\}}  
\end{split}
\end{equation}

Here, the first component calculates the cost based on the distance between two pedestrians' trajectories at each time point $t$, factoring in time discounting. The second term signifies whether pedestrian $i$'s trajectory intersects with a static obstacle in geometrically restricted environments. $\psi$ is not symmetric, meaning $\psi(\tau^{(i)},\tau^{(j)}) \neq \psi(\tau^{(j)},\tau^{(i)})$, due to the second term.

\subsubsection{Sampling-Based Opimization}
\paragraph{Initialization}

We use the non-interference distribution as the initial value for the peer distribution:
\begin{equation}
    \{\boldsymbol{\tau}_{i}\}^{(0)}_m = \{\boldsymbol{\tau}_{i}\}^{(NID)}_m = \{  \boldsymbol{\tau}_{i,1}^{(0)}, \boldsymbol{\tau}_{i,2}^{(0)}, ..., \boldsymbol{\tau}_{i,m}^{(0)} \}
\end{equation}
Where $\boldsymbol{\tau}_{i,j}^{(0)} = (\tau_{i,j}, w_{(i,j)}^{(0)})$. The optimization process solely focuses on refining the weights associated with the trajectories, while maintaining the inherent shape and structure of the trajectories unaltered.

\paragraph{Update}
Given the high-dimensional spaces' complexity, Monte Carlo integration is used to estimate the integrals, updating the sample's weight as:
\begin{equation}
\begin{split}
\label{eq:peer_inter_cost}
    \overline{\gamma_i}^{(l)}(\tau_{i,y}) &= \sum_{j=1}^{i-1}\frac{1}{m}\sum_{z=1}^m\psi(\tau_{i,y}, \tau_{j,z})w_{j,z}^{(l+1)} \\
    &+ \sum_{j=i+1}^{n}\frac{1}{m}\sum_{z=1}^m\psi(\tau_{i,y}, \tau_{j,z})w_{j,z}^{(l)}
\end{split}
\end{equation}

The weight of the sample can be updated as:
\begin{equation}
w_{i,j}^{(l+1)} = w_{i,j}^{(l)} \exp{-\overline{\gamma_i}^{(l)}(\tau_{i,y})}
\end{equation}

After updating all agents' sampled weights, we normalize the weights to keep the average weight at 1. For each $ \boldsymbol{\tau}_{i,j}^{(0)} = (\tau{i,j}, w_{(i,j)}^{(0)})$, the updated trajectory is approximated as:
\begin{equation}
    p_i^{PID}(\tau) = p_i^{(l)}(\tau_{i,j}) \approx w_{i,j}^{(l)}p_i^{(0)}(\tau_{i,j})
\end{equation}
% \paragraph{Pedestrian Trajectory Distribution Estimation}
% For computational convenience, we use a deterministic expression to estimate the pedestrian trajectory distribution:
% \begin{equation}
% p_i^{peer-deter}(\tau_{i}) = \delta(j - \arg max_j p_i^{(k)}(\tau_{i,j})) *  p_i^{(k)}(\tau_{i,j})
% \end{equation}

\subsection{Total-Influence Distribution}
Our analysis culminates in a scenario where pedestrians are aware of the robot's trajectory, introducing the 'Total-Influence Distribution'. This distribution incorporates the robot's trajectory as a dynamic obstacle in the pedestrians' cost function, allowing us to observe how pedestrians adapt their trajectories considering both the movements of other pedestrians and the robot.

The influence of the robot on pedestrian preference distributions is not unidimensional, but rather significantly varies based on the robot's specific actions. For instance, if a robot obstructs a pedestrian's direct path, the pedestrian would need to deviate. A slight move to the right might require minor adjustments from the pedestrian, while creating ample space could allow pedestrians to maintain their original path.
Given this variability in potential robot actions and their impacts, it becomes essential to consider each possible robot trajectory individually. This approach differs from the one proposed in \cite{sun2021move}, where the robot is treated as an additional pedestrian, and the weight of the entire trajectory is optimized within the optimization equation. In contrast, our method extracts only a single robot trajectory at a time for modeling the pedestrian's response.
This approach allows us to analyze how pedestrians might react under the influence of each specific robot trajectory. Particularly, it highlights the differences in pedestrian responses when a robot executes a specific trajectory versus when no robot is present. These differences are extracted to serve as our criteria for modeling micro-level intrusiveness.

In the following sections, we will delve into how we model the reactions of pedestrians using the multi-agent game-theoretic approach when the robot follows a specific trajectory. We consider a total of K possible trajectories $\{\tau_{Robot, 1}, \tau_{Robot,2}, ..., \tau_{Robot,K} \}$ for the robot. Without loss of generality, we focus our discussion on the modeling of pedestrian responses when the robot follows a specific trajectory, denoted as $\tau_{Robot, k}$, where $k \in \{1,2,...,K\}$.

 We consider the impact of the robot's trajectory on pedestrian interactions. Specifically, the robot's trajectory is factored into the calculation of the trajectory overlap penalty, as shown in the following equation:
\begin{equation}
\label{eq:aware_phi}
\begin{split}
      \psi(\tau^{(i)},\tau^{(j)}, & \tau_{Robot, k}) \\
      &= c_{ped} \max_{t} \gamma^t \cdot e^{\{-b \cdot [d(\tau^{(i)}(t), \tau^{(j)}(t)) - th_{peer}]  \}} \\
      &+ c_{obs}( \cdot \mathbbm{1}_{\{{\tau^{(i)} \cap M_{obs} \neq \emptyset}\}}  \\
      &+ \cdot \mathbbm{1}_{\{ 
       \min_{t}{d(\tau^{(i)}(t), \tau_{Robot, k}(t)) < th_{robot}}\}} \cdot \mathbbm{1}_{aware})
\end{split}
\end{equation}
% In our penalty framework, we include an awareness metric $\mathbbm{1}_{aware}$ to evaluate if the pedestrian $i$ is cognizant of robots, affecting his trajectory planning.
Our penalty model integrates an awareness metric $\mathbbm{1}_{aware}$, reflecting that unobservant pedestrians do not factor in robot avoidance in their trajectory planning. Further elaboration will be provided in Section \ref{sec:indepth_methodology}.
% \begin{equation}
% \begin{split}
%       \psi(\tau^{(i)},\tau^{(j)}, & \tau_{Robot, k}) \\
%       &= c_{ped} \max_{t} \gamma^t \cdot e^{\{-b \cdot [d(\tau^{(i)}(t), \tau^{(j)}(t)) - th_{peer}]  \}} \\
%       &+ c_{obs} \cdot \mathbbm{1}_{\{{\tau^{(i)} \cap M_{obs} \neq \emptyset}\}}  \\
%       &+ c_{obs} \cdot \mathbbm{1}_{\{ 
%        \min_{t}{d(\tau^{(i)}(t), \tau_{Robot, k}(t)) < th_{robot}}\}} \cdot \mathbbm{1}_{aware}
% \end{split}
% \end{equation}
\subsubsection{Optimization}
We initialize the process with the trajectories from the 'No Interference Distribution':
\begin{equation}
    \{\boldsymbol{\tau}_{i}\}^{(0)}_m = \{\boldsymbol{\tau}_{i}\}^{(NID)}_m  = \{  \boldsymbol{\tau}_{i,1}^{(0)}, \boldsymbol{\tau}_{i,2}^{(0)}, ..., \boldsymbol{\tau}_{i,m}^{(0)} \}
\end{equation}
Then, we calculate the pedestrian's reaction costs considering the $k$-th robot's trajectory:
\begin{equation}
\begin{split}
\label{eq:total_inter_cost}
    \overline{\gamma_i}^{(l) }(\tau_{i,y}| \tau_{Robot, k}) &= \sum_{j=1}^{i-1}\frac{1}{m}\sum_{z=1}^m\psi(\tau_{i,y}, \tau_{j,z},\tau_{Robot, k})w_{j,z}^{(l+1)} \\
    &+ \sum_{j=i+1}^{n}\frac{1}{m}\sum_{z=1}^m\psi(\tau_{i,y}, \tau_{j,z},\tau_{Robot, k})w_{j,z}^{(l)}
\end{split}
\end{equation}
The weight of each sample is updated based on the calculated cost:
\begin{equation}
w_{i,j}^{(l+1)} = w_{i,j}^{(l)} \exp{-\overline{\gamma_i}^{(l)}(\tau_{i,y}|\tau_{Robot, k})}
\end{equation}
For each $ \boldsymbol{\tau}_{i,j} = (\tau_{i,j}, w_{(i,j)})$, the updated for the trajectory is approximated as:
\begin{equation}
    p_i^{TID}(\tau|\tau_{Robot, k}) =p_i^{(l)}(\tau_{i,j}) \approx w_{i,j}^{(l)}p_i^{(0)}(\tau_{i,j})
\end{equation}
% \paragraph{Pedestrian Trajectory Distribution Estimation}
% For the sake of computational convenience, we utilize a deterministic expression.
% \begin{equation}
% p_i^{total-deter}(\tau_{i}|\tau_{k}^{Robot}) = \delta(j - \arg max_j p_i^{(k)}(\tau_{i,j})) *  p_i^{(k)}(\tau_{i,j})
% \end{equation}
Algorithm \ref{alg:individual_reaction} outlines the modeling of the distribution of disturbed trajectories for pedestrians.
When we set the robot's trajectory to be empty $\phi$, it models the peer-disturbance distribution for all individuals. When we set the robot's trajectory to be a specific valid trajectory from the sampled trajectories, it models the total-influence distribution corresponding to that particular behavior of the robot.

\subsection{Individual Disturbance Cost}
Upon calculating the 'Peer-Influence Distribution' and 'Total-Influence Distribution', we seek to quantify each individual pedestrian's behavioral changes due to the robot executing a specific trajectory, denoted as $\tau_{Robot, k}$. This quantification is achieved by calculating the Wasserstein Distance between the two distributions for each pedestrian.

% The Wasserstein distance adeptly quantifies the spatial discrepancy in pedestrian trajectories by measuring the exact cost of adjusting paths in response to a robot, offering a metric inherently attuned to the nuanced changes in pedestrian behavior and trajectory distribution shifts.
% The Wasserstein Distance, denoted as $W_1(P_1, P_2)$, excels in capturing the discrepancy between future trajectory distributions of pedestrians by directly quantifying the geometric shift in mass, making it intrinsically aligned with the spatial dynamics of pedestrian behavior in the presence of a robot.
% provides a meaningful geometric interpretation of the cost to transform one distribution into another, effectively capturing the nuances of multi-modal distributions and the sensitivity to the actual displacement of mass, making it a powerful metric for comparing distributions in a space where location and geometry are important. 
The Wasserstein distance is mathematically defined as follows:

\begin{equation}
\begin{split}
    &W_i(P_1, P_2) = \inf_{\gamma \in \Gamma(P_1, P_2)} \int_{\mathbb{R}^2} d(x, y) d\gamma(x, y)
    \\
    &\Gamma(P_1, P_2) = {\gamma(x,y) \mid \gamma(x, \cdot) = P_1(x), \gamma(\cdot, y) = P_2(y)}
\end{split}
\end{equation}
This metric adeptly quantifies the spatial discrepancy in pedestrian trajectories by measuring the exact cost of adjusting paths in response to a robot, offering a metric inherently attuned to the nuanced changes in pedestrian behavior and trajectory distribution shifts.
Specifically, $W_1(P_1, P_2)$ represents the 'shift' in behavior for a specific pedestrian when the robot is present and follows trajectory $\tau_{Robot, k}$, compared to when the robot is not present. A larger distance indicates that the robot's specific trajectory has a greater influence on the pedestrian's behavior, serving as an individual-specific measure of the robot's intrusiveness.

Given the Monte Carlo sampling method we used for optimization, we can express the Wasserstein Distance in a discrete form to accommodate our computation. This allows us to directly apply it to the sampled distributions of the pedestrian $i$:

\begin{equation}
\label{eq:standard_w_distance}
    W_i(P_1, P_2) = \min_{\Gamma \in \Pi(P_1, P_2)} \sum_{j=1}^{m} \sum_{j'=1}^{m} d(\tau_{i,j'}, \tau_{i, j'}) \cdot \gamma_{jj'}
\end{equation}

Where:

\begin{equation}
\Pi(P_1, P_2) = {\Gamma = {\gamma_{jj'}} \mid \gamma_{jj'} \geq 0, \sum_{j=1}^{n} \gamma_{jj'} = p_1^{(j)}, \sum_{j=1}^{m} \gamma_{jj'} = p_2^{(j)} }
\end{equation}

To simplify the calculations, we will convert the probability distribution of the stochastic model into a deterministic model before performing computations.
\begin{equation}
\begin{split}
    &W_i(\mathcal{F}(p_i^{TID}(\tau|\tau_{Robot, k})),\mathcal{F}(p_i^{PID}(\tau))) = \\
    &\sum_{j=1}^{m} \sum_{j'=1}^{m} d(\tau_j, \tau_{j'}) \cdot |\mathcal{F}(p_i^{TID}(\tau_j|\tau_{Robot, k})) - \mathcal{F}(p_i^{PID}(\tau_{j'}))|
\end{split}
\end{equation}

Here, $F(p) = \delta(j - argmax_j p_i(\tau_{i,j}))$ represents the transformation equation from a stochastic model to a deterministic model.

Given the scenario where the robot follows trajectory $k$, we denote the level of influence on pedestrian $i$ as $W_i(\mathcal{F}(p_i^{TID}(\tau|\tau_{Robot, k})),\mathcal{F}(p_i^{PID}(\tau)))$. In this context, individual disturbance is defined as the influence experienced by the most affected individual. Specifically, if we have a set of influence levels , where n is the total number of pedestrians, then the individual disturbance can be defined as:

\begin{equation}
    IDP = \max_i W_i(\mathcal{F}(p_i^{TID}(\tau|\tau_{Robot, k})),\mathcal{F}(p_i^{PID}(\tau)))
\end{equation}

Here, IDP represents individual disturbance, and max{} is the operation to select the maximum value from its elements. This allows us to quantify the maximum disturbance a pedestrian may experience given a certain robot trajectory.
Algorithm \ref{alg:idp} outlines outlines the detailed procedure for modeling individual disturbance.

\section{Flow Disturbance Modeling}
\label{sec:flow_disturbance_modeling}
\begin{figure}[htbp]
    \centering
    \includegraphics[width=\columnwidth]{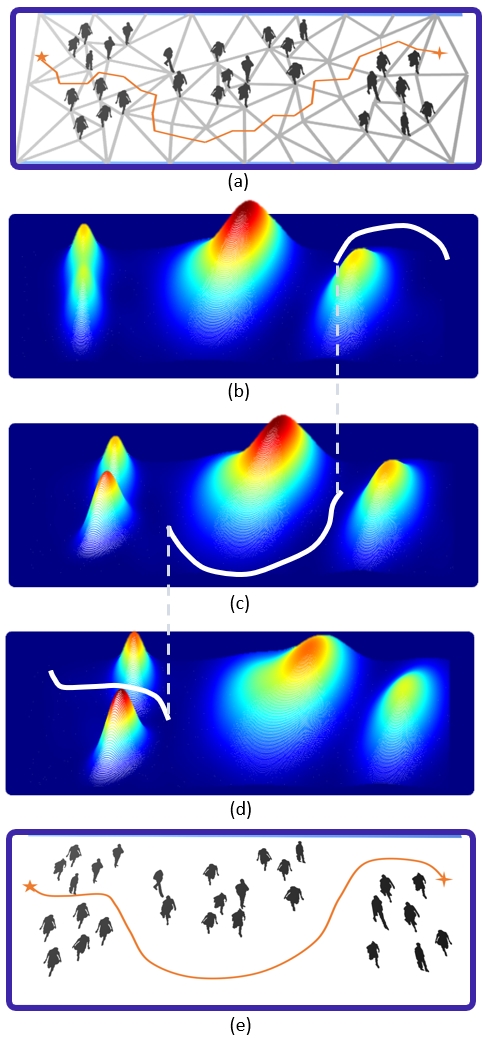}
    \caption{The process of Flow Disturbance Modeling. (a) The robot generates an initial estimate of the trajectory using a triangle-based search. (b, c, d) These time intervals represent variations in pedestrian flow, and the corresponding trajectories are calculated to evaluate the flow disturbance at each interval. (e) The optimized trajectory, obtained through optimization techniques, serves as a measure of the trajectory's impact on the flow and is considered the final assessment of flow disturbance within the framework.}
    \label{fig:fdp}
\end{figure}

In dense crowd navigation, due to the complexity of the environment, the estimation of long-term interactions is no longer accurate. Therefore, instead of excessively pursuing individual interactions in long-term planning, we focus more on the overall impact of the robot's decisions on the entire flow.

\subsection{Mathematical Formulation}
We introduce "flow disturbance," a measure of a robot's influence on human flow flux. This concept is defined by comparing the flux when a robot is present versus when it is absent, as formalized below:

\begin{equation}
\begin{split}
&\int_0^T \int_{R_d(t, R(t))} |\rho(t,x, R(t)) v(t, x, R(t)) - \rho_0(t,x) v_0(t, x)| dx dt \\
&=\int_0^T \int_{R_d(t, R(t))} | - \rho_0(t,x) v_0(t, x)| dx dt
\end{split}
\end{equation}
Here, the flux within the robot's envelope contour $R_d(t, R(t))$ is examined. Without a robot, the flux is represented by the product of crowd density ($\rho_0$) and velocity ($v_0$). With a robot, the flux within the envelope $\rho(t,x, R(t)) v(t, x, R(t))$ falls to zero due to the robot occupying the human flow's space.
This model focuses on macroscopic changes in the flow due to the robot's presence, negating the need for detailed microscopic interaction modeling.
We assume the crowd yields the robot's envelope space, simplifying the model.
Notably, the robot's position—and therefore its envelope—changes over time, influenced by the robot's decisions. Our model accounts for these dynamic changes in human flow and the robot's decisions.

 Accorgin to \cite{bellomo2008modelling}, the density and velocity of pedestrian flow can be represented as:

\begin{equation}
	\begin{aligned}
		\rho({\mathbf{x}},t) &= \sum_{j\in \mathcal{R}}f(\mathbf{r}_j(t)-\mathbf{x}(t)) \\
		&= \sum_{j\in \mathcal{R}} \frac{1}{\pi R^2}\exp(-\|\mathbf{r}_j(t)-\mathbf{x}(t) \|^2),
	\end{aligned}
\end{equation}

\begin{equation}
	v(\mathbf{x},t)=\frac{\Sigma_{i\in \mathcal{R}} \mathbf{v}_if(\mathbf{r}_i(t)-\mathbf{x}(t))}{\Sigma_{j\in \mathcal{R}}f(\mathbf{r}_j(t)-\mathbf{x}(t))},
 \end{equation}
where ${\mathbf{r}}_j(t)$ is the position of pedestrian $j$ at time $t$ within a specific region $\mathcal{R}$ that takes $\boldsymbol{r}$ as centers and $R$ as the radius.

In the absence of a robot, the model of pedestrian flow can be expressed using the following equations, derived from the research in \cite{bellomo2008modelling}:

\begin{equation}
\begin{split}
&\partial_t \rho + \nabla_x \cdot (\rho v) = 0 \\
&\partial_t v + (v\cdot \nabla_x)v =F(\rho,v)
\end{split}
\end{equation}

These two equations respectively represent the physical principles of mass conservation and momentum conservation. By applying these equations, we can compute the flux velocity and density values at various locations and time intervals. This provides an important numerical basis for us to evaluate the quality of robot paths in subsequent analyses.

When a robot is present, the mass conservation equation remains unchanged, while the momentum conservation equation needs to be adjusted to include the impact of the robot:

\begin{equation}
\begin{split}
&\partial_t \rho + \nabla_x \cdot (\rho v) = 0 \\
&\partial_t v + (v\cdot \nabla_x)v =F(\rho,v, R_d(t, R(t)))
\end{split}
\end{equation}

It's worth noting that the interaction model between the robot and pedestrian flow might involve some complex fluid mechanics concepts, such as boundary layer modeling and congestion modeling. However, due to our unique measurement method, the values obtained from the flow disturbance modeling are independent of the specific robot-pedestrian interaction model. This allows us to sidestep the complexity of the issue and focus on understanding and evaluating the impact of the robot from a macroscopic perspective.

\subsection{Flow Sensitive Triangle Graph Searching}
We utilize triangulation to divide the space into areas with topological attributes, providing a new, triangle-based method for our search process that no longer relies on the pixel level. This approach greatly simplifies the search process, as it's easier to manage and process search within relatively fewer triangles than within a large number of pixels. Moreover, when the search direction crosses two pedestrian nodes, it actually represents passing through two people, thus imparting more semantic information to the search process, which is unachievable in pixel-level searches.

In the triangulation-based search, we employ the A* algorithm. Unlike traditional searches that use Euclidean distance as a cost function, we introduce two additional costs sensitive to pedestrian flow: "resistance cost" and "lubrication cost". For the $j$th segment $\tau_j$ that connects two adjacent triangles, they can be computed using the following formulas:

\begin{equation}
\begin{split}
&\mathbf{RC}_j = \max(-\tau_j^T \cdot \rho(x,t) v(x,t), 0) \\
&\mathbf{LC}_j = |\tau_j \times \rho(x,t) v(x,t) |
\end{split}
\end{equation}

These two costs are designed to penalize behaviors that may disrupt pedestrian flow, such as counterflow and crossflow, while also encouraging the robot to avoid high-density areas. This cost calculation method makes our search process more accurate and efficient.

After completing the search, we obtain a series of polylines that represent a set of optimal paths considering pedestrian flow from the current point to the destination. 
As shown in Figure \ref{fig:fdp}(a), these polylines will serve as initial estimates in the subsequent optimization process.
\subsection{Back-End Path Optimization}

After applying the triangle-level search, we obtain a series of polyline paths. However, these initial paths can't guarantee compliance with motion constraints. To address this issue, we employ Gaussian Process (GP) optimization to refine these initial paths.

\subsubsection{Prior Distribution of Smooth Trajectory}
We first consider a Gaussian process generated by a linear time-varying stochastic differential equation (LTV-SDE):
\begin{equation}
    \dot{\xi(t)} = \mathbf{A}(t) \xi(t) + \mathbf{u}(t) + \mathbf{F}(t)w(t)
 \end{equation}
This is a differential equation describing a stochastic process, where $\xi(t)$ is the state-time on the path, $\mathbf{u}(t)$ is the external input (assumed to be zero in this paper), $\mathbf{A}(t)$ and $\mathbf{F}(t)$ are time-varying matrices, and $w(t)$ is a white noise process
 \begin{equation}
     w(t) \in \mathcal{GP}(0, \mathbf{Q_C}\delta(t-t'))
 \end{equation}
where $Q_C$ is the pwoer-spectral density matrix and $\delta(t-t')$ is the Dirac delta function. A general solution to the LTV-SDE is given by
\begin{equation}
\xi(t) = \mathbf{\Phi}(t,t_0)\xi(t_0) + \int_{t_0}^t{\mathbf{\Phi}(t,x)(u(x) + \mathbf{F}(x)\mathbf{w}(x))}dx
\end{equation}
where $\mathbf{\Phi}(t, x) = \exp\int_x^tA(x)dx$ is the transition matrix. The mean and covariance of the function $\xi(t)$ is as follows:
\begin{equation}
    \mathbf{\mu}(t) = \mathbb{E}[x(t)] = \mathbf{\Phi}(t,t_0) \mathbb{\mu}_0 + \int_{t_0}^t\mathbf{\Phi}(t,s)\mathbf{u}(s)ds
\end{equation} 
\begin{equation}
\begin{split}
        &\mathbf{\mathcal{K}}(t, t') = \mathbf{\Phi}(t,t_0)\mathcal{K}_0\mathbf{\Phi}(t',t_0)^T + \\
        &\int_{t_0}^{\min(t, t')}\mathbf{\Phi}(t,s)\mathbf{F}(s)Q_c\mathbf{F}(s)^T\mathbf{\Phi}(t',s)^Tds
\end{split}
\end{equation} 
With this mean and covariance, we can define the prior distribution of the Gaussian process:
\begin{equation}
    P(\xi) \propto  \exp\{ -0.5 \| \xi - \mathbf{\mu}  \|^2_{\mathcal{K}}\}
\end{equation}
Suppose a near-optimal time assignement, $t_0 < t_1, ..., t_i, < ..., <t_F, t_i = t_0 + d_t * i$, is found by the graph searching, then $\xi$ is discretized into $F+1$ waypoints accordingly. In general, each waypoint $\xi(t_i)$ is correlated with others, and the covariance matrix $\mathcal{K} \in \mathbb{R}^{F \times F}$ is dense. However, it is proved that LTV-SDE is with Markov property, and $\mathcal{K}$ has a very sparse structure. 
\begin{equation}
\begin{split}
    &P(\xi) \propto  \exp\{ -0.5 \sum_i e_i^T \mathbf{Q}_i^{-1}e_i\} \\
    & e_i = u(t_i) - \xi(t_i) + \mathbf{\xi}(t_i,t_{i-1}) \xi(t_{i-1}) \\
    & \mathbf{Q}_i = \int_{t_i -1}^{t_i} \mathbf{\Phi}(t,x)\mathbf{F}(x)Q_c\mathbf{F}(x)^T\mathbf{\Phi}(t',x)^Tdx
\end{split}
\end{equation}
which actually measures the difference between the actuval state $\xi(t_i)$ and $ u(t_i) + \mathbf{\xi}(t_i,t_{i-1}) \xi(t_{i-1})$ by the system w.r.t $\mathbf{Q}_i$. The smaller the difference is, the less system changes or control efforts are needed, and thus the smoother the trajectory is.
% \subsubsection{Timed-ESDF}
\subsubsection{Temporal Pedestrian Flow Influence Field}
Through Temporal Pedestrian Flow Influence Field (PFIF), flow influence and gradient queries can be constructed using state-time keys.
Incorporating flow information with a time dimension, we define a likelihood function to evaluate the impact exerted by a path:

\begin{equation}
p(\xi(t_i)) \in \mathcal(AST)|\xi(t_i) \propto \exp\{ -0.5 | \mathbf{h}(\xi(t_i)) |^2_{\sum_i}\}
\end{equation}

The function $\mathbf{h}(\xi(t_i))$ is derived by integrating the absolute difference between the robot's waypoint  $\xi(t)$ velocity component and the pedestrian flow velocity $v_0(t, x)$. This integration is performed across all time instances $t$ and all spatial coordinates $x$ within the robot's envelope $R_d(t, R(t))$. Furthermore, each of these velocity differences is weighted by the corresponding pedestrian density $\rho_0(t,x)$:

\begin{equation}
\label{eq:flow_disturbance}
\mathbf{h}(\xi(t_i)) = \int_{t_{i-1}}^{t_i} \int_{R_d(t, R(t))} \rho_0(t,x) |(\xi_v(t)-v_0(t, x))| dx dt
\end{equation}

Where $\xi_v(t)$ is the velocity component of $\xi(t)$, and $\mathbf{h}(\xi(t_i))$ provides a quantified measure of the impact of the path $\xi(t_i)$ on the pedestrian flow, considering both the magnitude of discrepancy between the robot's velocity and the flow velocity, as well as the pedestrian density.

Figure \ref{fig:fdp}(b-d) illustrates the steps involved in constructing the temporal pedestrian flow influence field. Firstly, based on fluid dynamics, we can calculate a spatio-temporal map of pedestrian flow in the absence of robots, using the current flow field information. Then, at different time intervals, we select different temporal maps to measure the density and velocity of pedestrian flow at a given waypoint location. Through optimization methods, we can obtain a smooth path with minimal disturbance, as shown in \ref{fig:fdp}(e).

\subsubsection{MAP Trajectory}
With the Gaussian Process (GP) prior and the likelihood function in place, we complete the back-end trajectory optimization by solving a Maximum A Posteriori Probability (MAP) problem. This optimization problem can be formulated as follows:
\begin{equation}
\label{eq:map}
\begin{split}
    \xi^* = & arg \max_{\xi}\{ p(\xi(t)) \prod_i  p(\xi(t_i) \in \mathcal(AST))  \} \\ 
    & arg \min_{\xi}\{  -log(p(\xi(t)) \prod_i  p(\xi(t_i) \in \mathcal(AST)) ) \} \\ 
    & arg \min_{\xi}\{ 0.5 \| \xi - \mathbf{\mu}  \|^2_{\mathcal{K}} + 0.5 \| \xi - \mathbf{h}(\xi(t_i)) \|^2_{\sum_i} \}
\end{split}
\end{equation}
Numerous numerical optimizers exist for solving Eq. (17), such as the Gauss-Newton or Levenberg-Marquardt optimizers. However, it's important to note that the accuracy and success rate of these numerical methods significantly depend on the quality of initial values. This highlights the necessity of a front-end path searching phase to provide good initial values for successful and accurate optimization. The entire algorithm's workflow is outlined in Algorithm \ref{alg:fdp}.

\begin{algorithm}
\caption{Individual Reaction Modeling (IRM)}\label{alg:individual_reaction}
\begin{algorithmic}[1]
\State \textbf{Input:}
\State \hspace{\algorithmicindent} $[\{\tau_{1}\}, \{\tau_{2}\}, ..., \{\tau_{n}\}]$: Samples of potential response trajectories of n pedestrians, each pedestrian has m samples, $    \{\tau_{i}\} = \{\tau_{i,1}, \tau_{i,2}, ..., \tau_{i,m} \}$. 
\State \hspace{\algorithmicindent} $\tau_{(Robot, k})$: the $k$-th sample trajectory of the robot. When we disregard the presence of the robot in peer disturbance modeling, we set it as $\phi$. When we consider the robot executing trajectory k in total disturbance modeling, we set it a valid trajectory. 
\State \hspace{\algorithmicindent} $[p_1(\boldsymbol{\tau}), p_2(\boldsymbol{\tau}), ..., p_n(\boldsymbol{\tau})]$: Original trajectory distribution of each pedestrian. 
\State \hspace{\algorithmicindent} $\mathcal{M}$: Map representation providing geometry-constrained terrain information for static obstacle collision detection.
\State \hspace{\algorithmicindent} $\epsilon$: Termination conditions.
\State \textbf{Output:}
\State \hspace{\algorithmicindent} $[p_1^{\textquotesingle}(\boldsymbol{\tau}), p_2^{\textquotesingle}(\boldsymbol{\tau}), ..., p_n^{\textquotesingle}(\boldsymbol{\tau})]$: Disturbed distribution, can be either peer-influenced distribution or total-influenced distribution.

\For{$i \gets 1$ to $n$}
    \For{$j \gets 1$ to $m$}
        \State $w_{i,j} \gets 1$
        \State $\boldsymbol{\tau}_{i,j} = ({\tau}_{i,j}, w_{i,j})$
     \EndFor

     $\{\boldsymbol{\tau}_{i}\}^{(0)}_m = \{  \boldsymbol{\tau}_{i,1}, \boldsymbol{\tau}_{i,2}, ..., \boldsymbol{\tau}_{i,m} \}$
\EndFor
\State $l \gets 0$
\coloredCommentTwo{Joint Expected Collision Penalty Calculation}
\State $Jc \gets \text{Eq. }($\ref{eq:joint_expected_collision_penalty}$)$
\While{$J_c > \epsilon$}

    \For{$i \gets 1$ to $n$}
        \For{$j \gets 1$ to $m$}

            \If{$\tau_{(Robot, k)}$ is $\phi$}

                \coloredComment{for Peer Disturbance Modeling}
            
                \State $v \gets Eq. ($\ref{eq:peer_inter_cost}$)$
            \Else
                \coloredComment{for Total Disturbance Modeling}
                \State $v \gets Eq. ($\ref{eq:total_inter_cost}$)$
            \EndIf
            \coloredComment{Solution to Iterative Best Response}
            \State $w_{i,j}^{(l+1)} \gets w_{i,j}^{(l)} \cdot \exp(-v/m)$
         \EndFor
        \For{$j \gets 1$ to $m$}       
            \State $w_{i,j}^{(l+1)} \gets w_{i,j}^{(l+1)} / (\sum_{ll=0}^m w_{i,j}^{(l+1)}/m)$
         \EndFor
    \EndFor
    \coloredCommentTwo{Joint Expected Collision Penalty Update}
    \State $Jc \gets \text{Eq.}
    ($\ref{eq:joint_expected_collision_penalty}$)$
\State $l \gets l+1$
\EndWhile

\For{$i \gets 1$ to $n$}
    \For{$j \gets 1$ to $m$}
        \State $ p_i^{\textquotesingle}(\tau_{i,j}) = w_{i,j}^{(l)}p_i^{(0)}(\tau_{i,j})$
     \EndFor
\EndFor

\State \textbf{return} $[p_1^{\textquotesingle}(\boldsymbol{\tau}), p_2^{\textquotesingle}(\boldsymbol{\tau}), ..., p_n^{\textquotesingle}(\boldsymbol{\tau})]$

\end{algorithmic}
\end{algorithm}

\begin{algorithm}
\caption{Individual Disturbance Modeling}\label{alg:idp}
\begin{algorithmic}[1]

\State \textbf{Input:}
% \State \hspace{\algorithmicindent} $[\{\tau_{1}\}, \{\tau_{2}\}, ..., \{\tau_{n}\}]$: Samples of potential response trajectories of n pedestrians, each pedestrian has m samples, $    \{\boldsymbol{\tau}_{i}\} = \{{\tau}_{i,1}, {\tau}_{i,2}, ..., {\tau}_{i,m} \}$. 

\State \hspace{\algorithmicindent} $\tau_{(Robot, k})$: the $k$-th sample trajectory of the robot.
\State \hspace{\algorithmicindent} $ped_1, ped_2, .., ped_n$: n pedestrians
\State \hspace{\algorithmicindent} $\mathcal{M}$: Map representation
% \State \hspace{\algorithmicindent} $[p_1(\boldsymbol{\tau}), p_2(\boldsymbol{\tau}), ..., p_n(\boldsymbol{\tau})]$: Original trajectory distribution of each pedestrian. 

% \State \hspace{\algorithmicindent} $\epsilon$: Termination conditions.
\State \textbf{Output:}
\State \hspace{\algorithmicindent} $IDP$: Individual Disturbance Penalty

\For{$i \gets 1$ to $n$}
    \For{$j \gets 1$ to $m$}
    \coloredCommentThree{Single Trajectory and Its Probability}
    \State $(\tau_{i,j}, p(\tau_{i,j})) \sim P_i^{NID}(\tau)$
    \EndFor
    \coloredCommentThree{Response Trajectories of Pedestrian i}
    \State $\{\tau_{i}\} \gets \{\tau_{i,1}, \tau_{i,2}, ..., \tau_{i,m} \}$
    \coloredCommentThree{Non Interference Distribution of  Pedestrian i}
    \State $p_i \gets \{p(\tau_{i,1}), p(\tau_{i,2}), ... p(\tau_{i,m})\}$
\EndFor

    \coloredCommentTwo{Peer Interference Distribution}
  \State $[p_1^{PID}, ..., p_n^{PID}] \gets Individual Reaction Modeling($
  \Statex \hspace{\algorithmicindent} $[\{\tau_{1}\}, \{\tau_{2}\}, ..., \{\tau_{n}\}], \phi, [p_1, p_2, ..., p_n] ,\mathcal{M}, \epsilon)$
  
\coloredCommentTwo{Total Interference Distribution}
  \State $[p_1^{TID}, ..., p_n^{TID}] \gets Individual Reaction Modeling($
  \Statex \hspace{\algorithmicindent} $[\{\tau_{1}\}, \{\tau_{2}\}, ..., \{\tau_{n}\}], \tau_{(Robot, k}), [p_1, p_2, ..., p_n] ,\mathcal{M}, \epsilon)$

\For{$i \gets 1$ to $n$}
    \coloredCommentThree{Wasserstein Distance between PID and TID of Pedestrian i}
    \State $W_i(p_i^{PID}, p_i^{TID}) \gets Eq. ($\ref{eq:standard_w_distance}$)$
\EndFor
\coloredCommentTwo{Max Wasserstein Distance among all Pedestrians}
\State $IDP \gets max_i W_i(p_i^{PID}, p_i^{TID})$
\State \textbf{return} $IDP$
% \State \hspace{\algorithmicindent} Index of $key$ in array $A$ if present, otherwise \textit{Not found}

\end{algorithmic}
\end{algorithm}

\begin{algorithm}
\caption{Flow Disturbance Modeling}
\label{alg:fdp}
\begin{algorithmic}[1]

\State \textbf{Input:}
\State \hspace{\algorithmicindent} $\tau_{(Robot, k})$: the $k$-th sample trajectory of the robot.
\State \hspace{\algorithmicindent} $\mathcal{M}$: Map representation, including the static map and spatio-temporal flow map of pedestrians.
\State \textbf{Output:}
\State \hspace{\algorithmicindent} $FDP$: Flow Disturbance Penalty

\State $start \gets \text{getFinalState}(\tau_{(Robot, k)})$ \Comment{The endpoint state of the partial trajectory is used to measure flow disturbance}

\State $\xi(t_0), \xi(t_1), ... \xi(t_F) \gets FlowSensitiveGraphSearching(\tau_{(Robot, k)}, goal)$
\State $\boldsymbol{\xi(t)}^{(0)} \gets \{ \xi(t_0), \xi(t_1), ... \xi(t_F) \}$

\State $f_{\text{prev}} \gets +\infty$ 
% \Comment{Previous best objective function value, initialized to infinity}
% \State $f_{\text{tol}} \gets \text{tolerance}$ \Comment{Convergence tolerance}
\State $l \gets 0$ 
% \Comment{Iteration counter}

\While{true}
    \State $\boldsymbol{\xi}^{(l+1)} \gets \text{UpdateParameters}(\boldsymbol{\xi}^{(l)})$ \Comment{Update parameters using optimization step  accroding to Eq. (}\ref{eq:map}$)$
    \State $f_{\text{new}} \gets \text{FlowDisturbance}(\boldsymbol{\xi}^{(l+1)})$ \Comment{Evaluate objective function according to Eq. (}\ref{eq:flow_disturbance}$)$
    \If{$|f_{\text{new}} - f_{\text{prev}}| < f_{\text{tol}}$} \Comment{Check for convergence}
        \State \textbf{break} \Comment{Exit the loop if the change is less than the tolerance}
    \EndIf
    \State $f_{\text{prev}} \gets f_{\text{new}}$ \Comment{Update the previous best value}
    \State $l \gets l + 1$ \Comment{Increment the iteration counter}
\EndWhile

\State $FDP \gets \text{FlowDisturbance}(\boldsymbol{\xi}^{(l+1)})$
%\State $FDP \gets Eq. ($\ref{eq:flow_disturbance}$)$
\State \textbf{return} $FDP$
% \State \hspace{\algorithmicindent} Index of $key$ in array $A$ if present, otherwise \textit{Not found}

\end{algorithmic}
\end{algorithm}

\section{In-depth Methodologies and Practical Execution}
\label{sec:indepth_methodology}

% \section{Flow Disturbance Modeling}
% \label{sec:flow_disturbance_modeling}
\begin{figure}[t]
    \centering
    \includegraphics[width=\columnwidth]{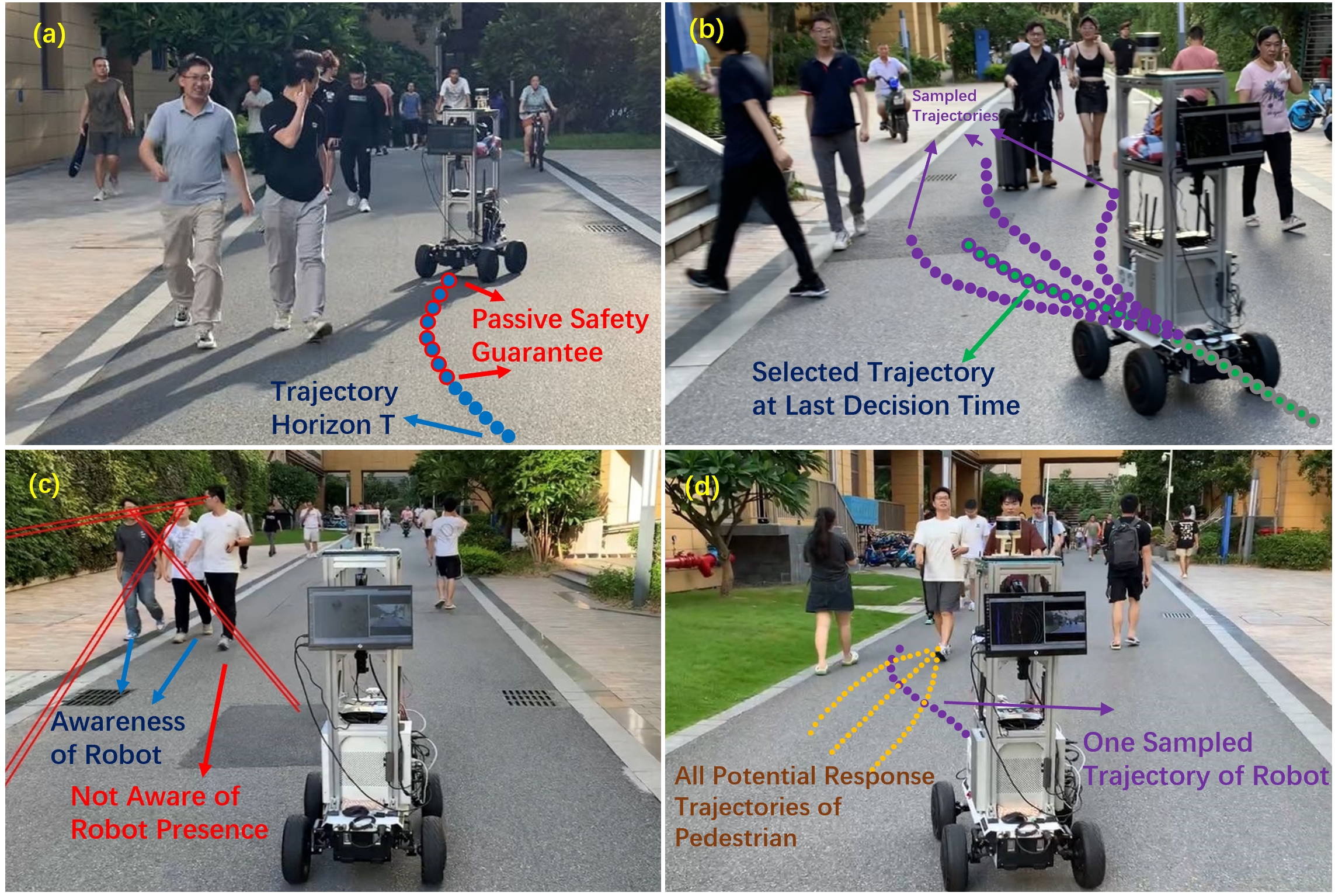}
    \caption{In-depth Methodologies and Practical Execution. (a) Passive safety guarantee. the blue waypoints with red edges, corresponds to the passive safety zone. (b) Temporal trajectory consistency mechanism. The green trajectory is the previous optimal choice, with gray edges for the executed portion and purple edges for the unexecuted portion. The solid purple trajectories are new samples. (c) Cognition of robotic presence among pedestrians accroding to gaze direction. (d) All potential pedestrian responses resulting in collisions with a specific robot trajectory prevent the optimization of the pedestrian's trajectory distribution.}
    \label{fig:in-depth}
\end{figure}

The deployment of autonomous robots within pedestrian milieus necessitates a nuanced consideration of safety protocols, the clarity of robotic trajectories, and the engagement of pedestrian awareness. The practical application of individual and flow disturbance models brings forth complexities that extend beyond theoretical constructs. To address these intricacies, our proposed framework encompasses a multipronged approach: instituting a passive safety paradigm to safeguard against the unforeseen variances in pedestrian behavior, implementing a temporal trajectory consistency mechanism to foster legible and stable robotic navigation, and enhancing detection methodologies to evaluate and predict pedestrian responsiveness to the robot's presence. These tailored solutions are pivotal in ensuring that the integration of autonomous systems into human-centric environments is not only theoretically sound but also practically viable and socially harmonious.

\subsection{Passive Safety Guarantee}
For effective navigation involving robots and pedestrians, the optimal circumstance would be for pedestrians to interact with the robot in a manner entirely consistent with the robot's predictions. Nevertheless, in cases where pedestrian behavior deviates from the robot's expectations, it's vital to maintain a high degree of safety.

In this context, we advocate for the principle of passive safety, which is realized through the avoidance of Inevitable Collision States (ICS). An ICS is characterized as a state that, once entered by the robot, will inevitably lead to a collision with obstacles within a specific time interval. Maintaining an ICS-free state is synonymous with ensuring passive safety, effectively preempting potential collisions by never entering such states, and tantamount to ensuring a full stop prior to a potential collision.

To elegantly capture the essence of individual disturbances while preserving safety, our strategy bifurcates the robot's trajectory into two segments. The initial portion of the trajectory, denoted as 'safe' zone, adheres to passive safety, where the robot's path is meticulously designed to avoid any Inevitable Collision States (ICS), thereby providing an unequivocal assurance of safety in the near-term portion of the trajectory.
In contrast, the latter segment of the trajectory, denoted as 'negotiable' zone, is crafted to allow for potential collisions. This is not to suggest a disregard for safety, but rather an intentional allowance for the dynamic negotiation between robot and pedestrian. It is within this space that the robot subtly communicates its intended path, permitting a degree of uncertainty that encourages pedestrian response and adaptation. This nuanced approach enables the modeling of a reciprocal and exploratory interaction, where the pedestrian and the robot engage in a tacit ballet of move and countermove.

In the context of passive safety verification, we have drawn upon our prior research methodology \cite{9560852}, which entails the comprehensive testing of all minimum-acceleration motion primitives at the designated endpoint. However, it is important to note that in the present task, the term "endpoint" does not merely denote the termination point of the trajectory, but rather signifies the culmination of the 'safety' zone.
As shown in Figure \ref{fig:in-depth} (a), The entire blue trajectory represents the overall horizon of the partial trajectory. The front half of the trajectory, indicated by the blue waypoints with red edges, corresponds to the passive safety zone.

% Safety verification involves ensuring that no intersection occurs between unpredictable future human states and the initial two seconds of a candidate trajectory. However, identifying braking ICS is a labor-intensive task. To address this challenge, we suggest testing all minimum-acceleration motion primitives originating from the horizon of the 'safty zone'. If at least one collision-free primitive exists, the trajectory's safety can be assured. In this scenario, we only need to verify the minimum-acceleration primitive for each steering angle.

% we employ the principle of passive safety, a form of motion safety attained by evading Inevitable Collision States (ICS) to ensure safety. ICS is characterized as a state that, once entered by the robot, will inevitably lead to a collision with obstacles within a specific time interval. This is tantamount to ensuring a full stop prior to a potential collision.

% Safety verification involves ensuring that no intersection occurs between unpredictable future human states and the initial two seconds of a candidate trajectory. However, identifying braking ICS is a labor-intensive task. To address this challenge, we suggest testing all minimum-acceleration motion primitives originating from the robot's position. If at least one collision-free primitive exists, the trajectory's safety can be assured. In this scenario, we only need to verify the minimum-acceleration primitive for each steering angle.

\subsection{Temporal Trajectory Consistency Mechanism}
The robustness of a robot's decision-making process plays a critical role in its interactions with pedestrians. A high decision frequency may prevent pedestrians from discerning a stable intention from the robot, thereby undermining the legibility of these interactions. This challenge is further exacerbated in practical implementations due to perception noise and delays, potentially leading to oscillatory behavior as the robot continuously alternates between different trajectories.

While reducing the decision frequency might mitigate this to an extent, it could compromise the robot's capacity to promptly react to unforeseen events. To resolve this, we suggest a temporal trajectory consistency mechanism that ensures a degree of continuity in the robot's trajectory over short timeframes. This mechanism enables the robot to respond aptly to sudden changes while maintaining predictability for pedestrians.

Specifically, we consider the previously selected optimal trajectory from the last decision (minus the executed portion) alongside the resampled trajectories as candidates for the current trajectory selection. When computing the individual distribution penalty (IDP), we assign an additional time-decay weight to the trajectory chosen in the prior decision.

If a new trajectory is chosen in the current decision, the weight count for the next iteration is reset. Conversely, if the trajectory selected in the current decision mirrors the optimal trajectory from the last iteration, the weight count for the next iteration increases by 1, resulting in a weight decrease. Once the weight decreases to a certain threshold, it will be disregarded as a candidate trajectory.
As shown in Figure \ref{fig:in-depth} (b),in the current decision, the unexecuted portion of the optimal trajectory from the previous decision (depicted as green waypoints with purple edges), along with the newly sampled trajectories (depicted as solid purple lines), are considered as the trajectory candidates that the robot can choose from.

This mechanism maintains high decision frequency while ensuring flexibility in trajectory selection. If a newly chosen trajectory's score marginally surpasses the previous one, the preceding decision is retained. However, in the event of an unexpected situation where the new trajectory's score significantly outperforms the preceding one (considering the additional weight), the robot can switch to a new decision intention, thereby ensuring its ability to adapt to sudden events.

\subsection{Cognition of Robotic Presence Among Pedestrians}
\label{subsec: cognition}
Pedestrian awareness of the robot's presence is a pivotal element in robotic decision-making. 
Situations where pedestrians may be oblivious to the robot's presence—such as pedestrians engaged with their phones or when the robot is behind pedestrians and out of their line of sight—impede interactive behavior, including fundamental obstacle avoidance.
Pedestrians' lack of awareness regarding the presence of the robot results in their failure to engage in avoidance behavior. This modeling of awareness, reflected by the indicator function $\mathbbm{1}_{aware}$ in Eq. \ref{eq:aware_phi}, introduces penalties for trajectory overlap and subsequently influences the distribution of trajectories for all individuals involved.

% This awareness primarily impacts the calculation of individual disturbance in our algorithm, particularly influencing the trajectory overlap penalty as specified in Eq. \ref{eq:aware_phi}. The management of such circumstances will be addressed in the subsequent subsection. Our primary focus in this part is modeling and evaluating pedestrian awareness of the robot's presence.

% Pedestrians' lack of awareness regarding the presence of the robot results in their failure to engage in avoidance behavior. This modeling of awareness, reflected by the indicator function in formula 21, introduces penalties for trajectory overlap and subsequently influences the distribution of trajectories for all individuals involved.

To assess pedestrian awareness, we primarily employ two indicators: the physical orientation of the body and eye gaze information.
% For physical orientation, we assume a range of 120 degrees and a distance of 10 meters within which pedestrians can perceive other entities.
For physical orientation, we assume that pedestrians can perceive other entities within a range of 90 degrees relative to their forward direction and within a distance of 10 meters.
Regarding eye gaze, we stipulate that the angle between the direction of the 3D eye gaze vector and the direction from the pedestrian to the robot should fall within 60 degrees. If the robot is within the union of these two ranges, we consider it to have been noticed by the pedestrian. In such instances, we anticipate a degree of cooperative interaction from the pedestrian in our decision-making process. Figure \ref{fig:in-depth} (c) exemplifies the use of eye gaze to assess pedestrian awareness of the robot. Among the three pedestrians walking together, two individuals have the robot within their central gaze angle, indicating awareness. In contrast, the third person's gaze is averted, suggesting a lack of awareness and the inability to evade the robot's trajectory in their response modeling.

\subsection{Collision Checking}
Our pipeline bifurcates collision checking into pre-processing and post-processing stages. In the pre-processing stage, any trajectory candidate that fails to guarantee passive safety or is excessively close to static obstacles is immediately discarded, bypassing the computation of individual disturbance. However, the potential for collisions persists in the post-processing stage, primarily due to two scenarios:

Firstly, if the robot executes a trajectory that remains unnoticed by pedestrians, as discussed in the subsection \ref{subsec: cognition}, the pedestrian's trajectory distribution may remain largely unchanged, leading to a potential collision with pedestrians if the trajectory is continuously followed.
Secondly, even if a pedestrian is cognizant of the robot's presence during a specific trajectory selection, all potential pedestrian responses may still culminate in a potential collision. This can infuse biases into multi-agent game optimization, leading to possible collisions between the robot and pedestrians.
This situation can be illustrated by referring to Figure \ref{fig:in-depth} (d). As per Equation \ref{eq:sivu}, the optimization of the pedestrian's distribution may not occur in such cases.
Consequently, even after computing the individual disturbance, it remains necessary to conduct collision detection for each trajectory to obtain results post bias-correction.

\section{EXPERIMENT}
\label{sec:experiment}

 % zheshidi 1 ge
\begin{table*}[tbhp]
\centering
\renewcommand{\arraystretch}{1.5}
\caption{three Scenarios under no counterflow situation.}
\begin{tabular}
{p{0.05\textwidth}p{0.10\textwidth}p{0.08\textwidth}p{0.07\textwidth}p{0.07\textwidth}p{0.07\textwidth}p{0.06\textwidth}p{0.06\textwidth}p{0.09\textwidth}p{0.06\textwidth}p{0.07\textwidth}}
\hline
\multirow{2}{*}{Scenario} & \multirow{2}{*}{Method} & \multicolumn{9}{c}{Metrics} \\
\cline{3-11}
& & Complete Ratio (\%)  $\uparrow$ & Success Rate (\%) $\uparrow$ & Timeout Rate (\%) $\downarrow$ & Collision Rate (\%)  $\downarrow$ & Freezing Num $\downarrow$ &Jerk  $\downarrow$ &Frontal Interact Num $\downarrow$ & Density $\downarrow$ &Execute Time (s) $\downarrow$\\
\hline
\multirow{7}{*}{NC-DT}
& ORCA           & 80.40  & 66.50 &  13.50 &  20.00 &  30  &  23.509  &  116 & 436.880 &  49.378 \\
& Dist Navi      & 88.30  & 68.00 &  20.00 &  15.00 &  1   &  23.647  &  90  & 350.660 & 46.976 \\
& DWA            & 81.25  & 70.00 &  0.00  &  30.00 &  2   &  23.704  &  62  & 371.938 & 44.406 \\
& Risk RRT       & 67.72  & 47.50 &  2.50  &  50.00 &  4  &  23.183   &  73  & 223.564 & 34.524 \\
& Dyn Channel    & 88.35  & 62.50 &  10.00 &  27.50 &  36 &  25.513  & 109  &  376.318 & 46.452 \\
& Crowd Planner  & 94.55  & 87.50 &  0.00  &  12.50 & 1   &  22.539  &  58  &  258.157 & 37.334 \\
& Ours           & 97.00  & 90.00 &  2.50  &  7.500 & 0   &  22.402  &  73  &  329.681 & 34.938 \\
\cline{1-11}
\multirow{7}{*}{NC-UT}
& ORCA           & 31.40  & 5.00 &  12.50 &  82.50 &  88   &  23.176  &  212  & 344.278 &  35.903 \\
& Dist Navi      & 82.50  & 42.50 & 22.50 &  35.00 &  1    &  24.033  &  289  & 463.978 &  48.824 \\
& DWA            & 77.25  & 52.50 & 2.50  &  45.00 &  8    & 24.056   &  201  & 439.337 &  44.884 \\
& Risk RRT       & 66.60  & 40.00 & 0.00  &  60.00 &  60   & 22.777   &  179  & 344.534 &  40.189 \\
& Dyn Channel    & 48.75  & 17.50 & 10.00 &  72.50 &  28   & 27.917   &  300  & 480.923 &  56.615 \\
& Crowd Planner  & 71.10  & 45.00 & 0.00 &  55.00 &    1   & 23.268   &  202  & 356.268 &  40.155 \\
& Ours           & 80.40  & 70.00 & 0.00 &  30.00 &   2    & 23.831   &  148  & 290.558 &  31.542 \\
\cline{1-11}
\multirow{7}{*}{FI-DT}
& ORCA           & 91.55  & 77.50 &  5.00 &  17.50  &  21  &  27.866  &  49  & 540.699 &  48.908 \\
& Dist Navi      & 87.75  & 57.50 &  20.00 &  22.50 &  0   &  29.037  &  47  & 564.216 & 56.513 \\
& DWA            & 80.25  & 62.50 &  0.00  &  37.50 &  1   &  28.760  &  36  & 569.109 & 53.308 \\
& Risk RRT       & 71.00  & 40.00 &  7.50  &  52.50 & 54  &  39.288   &  28  & 473.141 & 45.338 \\
& Dyn Channel    & 70.50  & 35.00 &  17.50 &  47.50 &  30 &  30.122   &  58  & 559.529 & 55.390 \\
& Crowd Planner  & 77.00  & 60.00 &  0.00  &  40.00 & 1   &  31.104   &  43  & 495.168 & 48.961 \\
& Ours           & 95.05  & 82.50 &  2.50  &  15.00 & 0   &  29.942   &  23  & 488.019 & 41.260 \\
\cline{1-11}
\multirow{7}{*}{FI-NT}
& ORCA           & 77.75  & 57.50 &  0.00 &  42.50 &  0    &  34.955  &  120  & 519.861 &  42.906 \\
& Dist Navi      & 85.90  & 50.00 & 25.00 &  25.00 &  0    &  30.244  &  162  & 731.082 &  57.403 \\
& DWA            & 63.40  & 30.00 & 7.50  &  62.50 &  0    & 35.615   &  117  & 768.159 &  58.991 \\
& Risk RRT       & 61.65  & 32.50 & 10.00 &  57.50 &  84   & 31.257   &  144  & 686.413 &  54.825 \\
& Dyn Channel    & 63.45  & 27.50 & 30.00 &  42.50 &  102  & 34.135   &  227  & 816.473 &  72.966 \\
& Crowd Planner  & 85.23  & 72.50 & 0.00 &  27.50  &   30  & 24.006   &  139  & 642.473 &  49.384 \\
& Ours           & 94.20  & 82.50 & 0.00 &  17.50  &   0   & 24.222   &  51   & 461.951 &  34.692 \\
\cline{1-11}

\multirow{7}{*}{BA-DT}
& ORCA           & 94.10  & 87.50 &  0.00 &  12.50  &  2  &  22.056  &  26  & 311.509 &  32.757 \\
& Dist Navi      & 84.50  & 55.00 &  20.00 &  25.00 &  0   &  21.769  &  10  & 390.840 & 43.716 \\
& DWA            & 84.75  & 67.50 &  10.00  &  22.50 &  0   &  22.201  &  32  & 476.855 & 43.837 \\
& Risk RRT       & 85.10  & 67.50 &  0.00  &  32.50 & 1  &  22.045   &  21  & 222.504 & 29.025 \\
& Dyn Channel    & 77.15  & 62.50 &  0.00 &  37.50 &  1 &  20.324   &  29  & 259.779 & 32.832 \\
& Crowd Planner  & 90.80  & 87.50 &  0.00  &  12.50 & 0   &  21.547   &  15  & 394.064 & 35.881 \\
& Ours           & 94.30  & 90.00 &  0.00  &  10.00 & 0   &  21.474   &  8  & 334.496 & 29.586 \\
\cline{1-11}
\multirow{7}{*}{BA-NT}
& ORCA           & 72.00  & 50.00 &  10.00 &  40.00 &  9    &  24.014  &  104  & 517.064 &  46.889 \\
& Dist Navi      & 66.15  & 47.50 & 5.00 &  47.50 &  0    &  23.295  &  83  & 504.686 &  45.864 \\
& DWA            & 75.55  & 52.50 & 10.00  &  37.50 &  5    & 21.389   &  99  & 525.308 & 46.300 \\
& Risk RRT       & 63.75  & 47.50 & 0.00 &  52.50 &  2  & 22.541   &  65  & 305.537 &  31.975 \\
& Dyn Channel    & 78.95  & 52.50 & 25.00 &  22.50 &  40  & 22.432   &  229  & 455.245 &  25.819 \\
& Crowd Planner  & 60.00  & 40.00 & 0.00 &  60.00  &   35  & 21.300   &  93  & 460.341 &  44.803 \\
& Ours           & 90.60  & 87.50 & 0.00 &  12.50  &   3   & 21.313   &  29   & 311.442 & 28.382 \\
\cline{1-11}

\hline
\end{tabular}
\label{tab:no_counter}
\end{table*}

 % zheshidi 2 ge
\begin{table*}[tbhp]
\centering
\renewcommand{\arraystretch}{1.5}
\caption{three Scenarios with counterflow situation.}
\begin{tabular}
{p{0.05\textwidth}p{0.10\textwidth}p{0.08\textwidth}p{0.07\textwidth}p{0.07\textwidth}p{0.07\textwidth}p{0.06\textwidth}p{0.06\textwidth}p{0.09\textwidth}p{0.06\textwidth}p{0.07\textwidth}}
\hline
\multirow{2}{*}{Scenario} & \multirow{2}{*}{Method} & \multicolumn{9}{c}{Metrics} \\
\cline{3-11}
& & Complete Ratio (\%)  $\uparrow$ & Success Rate (\%) $\uparrow$ & Timeout Rate (\%) $\downarrow$ & Collision Rate (\%)  $\downarrow$ & Freezing Num $\downarrow$ &Jerk  $\downarrow$ &Frontal Interact Num $\downarrow$ & Density $\downarrow$ &Execute Time (s) $\downarrow$\\
\hline
\multirow{7}{*}{NC-DT}
& ORCA           & 99.15  & 97.50 &  0.00 &  2.50 &  0  &  22.689  &  0    & 399.306     &  40.546 \\
& Dist Navi      & 89.88  & 62.50 & 20.00 &  17.50 & 0  & 21.890   & 0     & 351.593     & 46.587 \\
& DWA            & 85.45  & 67.50 & 5.00  &  27.50 & 4  & 22.578   & 0     & 411.398     & 43.554 \\
& Risk RRT       & 93.70  & 90.00 & 0.00  &  10.00 & 1  & 22.072   & 0     & 217.577     & 30.843 \\
& Dyn Channel    & 97.45  & 87.50 & 12.50 &  0.00  & 46 & 23.304   & 0    & 372.398      & 44.381 \\
& Crowd Planner  & 98.95  & 97.50 & 2.50  &  0.00  & 1  & 24.922   & 0    & 250.298      & 37.736 \\
& Ours           & 100.00 & 100.00& 0.00  & 0.00   & 0  & 22.340   & 0    & 306.039      & 32.397 \\
\cline{1-11}
\multirow{7}{*}{NC-UT}
& ORCA           & 33.45  & 12.50 & 15.00 &  72.50 &  54  &  49.775  &  191  & 528.662&  46.641 \\
& Dist Navi      & 72.25  & 37.50 & 15.00 &  47.50 &   2  &  24.844  &  233  & 492,489 & 50.110 \\
& DWA            & 89.15  & 77.50 & 2.50  &  20.00 &   6  &  23.802  &  162  & 420.617 & 40.101 \\
& Risk RRT       & 63.90  & 40.00 & 0.00  &  60.00 &   6  &  23.912  &  201  & 307.203 & 34.427 \\
& Dyn Channel    & 36.40  & 12.50 & 5.00  &  82.50 &   23 &  26.418  &  284  & 468.035 & 57.500 \\
& Crowd Planner  & 83.80  & 72.50 & 0.00  & 27.50  &   0  &  22.995  &  199  & 380.668 & 38,281 \\
& Ours           & 94.60  & 82.50 & 0.00  & 17.50  &   1  &  23.065  &  130  & 297.630 & 29.413 \\
\cline{1-11}

\multirow{7}{*}{FI-DT}
& ORCA           & 95.00  & 90.00 &  7.50 &  2.50 & 41 &  27.672  &  0    & 580.565     & 47.368 \\
& Dist Navi      & 96.70  & 67.50 & 27.50 &  5.00 & 0  & 27.693   & 0     & 617.518     & 56.246 \\
& DWA            & 86.00  & 70.00 & 5.00  & 25.00 & 0  & 29.639   & 0     & 562.557     & 52.209 \\
& Risk RRT       & 79.25  & 62.50 & 10.00 & 27.50 & 62 & 29.653   & 2     & 548.933     & 45.470 \\
& Dyn Channel    & 90.25  & 82.50 & 5.00  & 12.50 & 50 & 27.878   & 0     & 555.213     & 52.318 \\
& Crowd Planner  & 85.25  & 72.50 & 0.00  & 22.50 & 0  & 30.006   & 0     & 602.917     & 47.318 \\
& Ours           & 97.75  & 87.50 & 2.50  & 10.00 & 0  & 28.532   & 0     & 517.780     & 41.257 \\
\cline{1-11}
\multirow{7}{*}{FI-UT}
& ORCA           & 81.75  & 62.50 & 2.50  &  35.00 &  14  &  38.165  &  124  & 561.211&  42.752 \\
& Dist Navi      & 89.70  & 57.50 & 25.00 &  17.50 &   0  &  31.739  &  130  & 783.739 & 56.900 \\
& DWA            & 70.50  & 47.50 & 2.50  &  50.00 &   0  &  37.815  &  115  & 777.305 & 56.315 \\
& Risk RRT       & 72.25  & 57.50 & 27.50 &  15.00 &  161 &  31.668  &  115  & 847.114 & 56.304 \\
& Dyn Channel    & 51.30  & 10.00 & 20.00 &  70.00 &   87 &  40.670  &  204  & 898.704 & 73.402 \\
& Crowd Planner  & 84.20  & 70.00 & 0.00  &  30.00 &   56 &  22.799  &  146  & 688.147 & 52.292 \\
& Ours           & 99.00  & 95.00 & 0.00  &  5.00  &   0  &  35.515  &  43  & 467.041 & 33.707 \\
\cline{1-11}

\multirow{7}{*}{BA-DT}
& ORCA           & 97.00  & 95.00 &  0.00 &  5.00 & 0 &  21.256  &  0    & 319.860     & 31.920 \\
& Dist Navi      & 93.70  & 67.50 &  25.00 &  7.50 & 0  & 20.533   & 0     & 391.324     & 43.124 \\
& DWA            & 90.90  & 80.00 & 5.00  & 15.00 & 0  & 23.223   & 0     & 436.491     & 41.568 \\
& Risk RRT       & 80.60  & 75.00 &  0.00 & 25.00 & 0 & 22.109   &  0    & 254.210     & 31.067 \\
& Dyn Channel    & 85.90  & 67.50 & 0.00  & 32.50 & 0 & 21.311   & 1     & 257.632     & 31.085 \\
& Crowd Planner  & 89.30  & 85.00 & 0.00  & 15.00 & 0  & 21.355   & 0     & 427.704     & 35.915 \\
& Ours           & 93.00  & 90.00 & 0.00  & 10.00 & 0  & 21.586   & 0     & 332.376     & 28.118 \\
\cline{1-11}
\multirow{7}{*}{BA-UT}
& ORCA           & 67.00  & 45.00 & 0.00  &  55.00 &  0  &  22.075  &  106  & 372.075&  36.190 \\
& Dist Navi      & 68.50  & 50.00 & 7.50 &  42.50 &   0  &  23.043  &  64  & 553.504 & 44.959 \\
& DWA            & 76.00  & 52.50 & 12.50  &  35.00 &   2  &  21.908  &  101  & 559.737 & 45.894 \\
& Risk RRT       & 90.70  & 77.50 & 0.00 &  22.50 &  0 &  20.837  &  59  & 354.509 & 29.349 \\
& Dyn Channel    & 74.60  & 57.50 & 15.00 &  27.50 &   26 &  22.506  &  197  & 406.903 & 45.147 \\
& Crowd Planner  & 65.80  & 40.00 & 20.00  &  40.00 &   61 &  24.568  &  131  & 503.152 & 49.073 \\
& Ours           & 91.70  & 87.50 & 0.00  &  12.50  &   0  &  21.496  &  21  & 330.185 & 27.241 \\
\cline{1-11}

\hline
\end{tabular}
\label{tab:with_counter}
\end{table*}

 % zheshidi 3 ge
\begin{table*}[tbhp]
\centering
\renewcommand{\arraystretch}{1.5}
\caption{The performance of the Narrow Space.}
\begin{tabular}
{p{0.05\textwidth}p{0.10\textwidth}p{0.08\textwidth}p{0.07\textwidth}p{0.07\textwidth}p{0.07\textwidth}p{0.06\textwidth}p{0.06\textwidth}p{0.09\textwidth}p{0.06\textwidth}p{0.07\textwidth}}
\hline
\multirow{2}{*}{Density} & \multirow{2}{*}{Method} & \multicolumn{9}{c}{Metrics} \\
\cline{3-11}
& & Complete Ratio (\%)  $\uparrow$ & Success Rate (\%) $\uparrow$ & Timeout Rate (\%) $\downarrow$ & Collision Rate (\%)  $\downarrow$ & Freezing Num $\downarrow$ &Jerk  $\downarrow$ &Frontal Interact Num $\downarrow$ & Density $\downarrow$ &Execute Time (s) $\downarrow$\\
\hline
\multirow{7}{*}{Low}
& ORCA           & 98.53  & 92.50 &  2.50 &  5.00 &  0  &  20.847  &  26    & 475.544     &  35.429 \\
& Dist Navi      & 98.19  & 85.00 & 10.00 &  5.00 & 13  & 23.011   & 21     & 605.820     & 43.657 \\
& DWA            & 99.50  & 95.00 & 0.00  &  5.00 & 0  & 22.630   & 12     & 482.725     & 35.422 \\
& Risk RRT       & 93.15  & 87.50 & 0.00  &  12.50 & 3  & 23.069   & 29     & 410.129     & 38.026 \\
& Dyn Channel    & 96.31  & 85.00 & 5.00 &  10.00  & 12 & 22.968   & 56    & 529.001      & 39.636 \\
& Crowd Planner  & 100.00  & 100.00 & 0.00  &  0.00  & 3  & 21.995   & 26    & 485.483      & 34.017 \\
& Ours           & 100.00 & 100.00& 0.00  & 0.00   & 0  & 22.838   & 4    & 407.475      & 27.297 \\
\cline{1-11}
\multirow{7}{*}{High}
& ORCA           & 69.22  & 57.50 & 0.00 &  42.50 &  9  &  22.780  &  60  &  784.675&  41.133 \\
& Dist Navi      & 76.91  & 52.50 & 17.50 &  30.00 &   33  &  22.894  &  68  & 632.937 & 30.427 \\
& DWA            & 92.04  & 70.00 & 25.00  &  5.00 &   25  &  22.582  &  60  & 826.651 & 42.274 \\
& Risk RRT       & 72.68  & 65.00 & 25.00  &  10.00 &   50  &  23.464  &  61  & 1013.053 & 38.654 \\
& Dyn Channel    & 80.84  & 60.00 & 10.00  &  30.00 &   43 &  25.611  &  143  & 718.184 & 43.520 \\
& Crowd Planner  & 91.00  & 80.00 & 0.00  & 20.00  &   12  &  22.411  &  79  & 721.336  & 36.453 \\
& Ours           & 97.61  & 85.00 & 5.00  & 10.00  &   3  &  23.654  &  27   & 701.995 & 31.259 \\
\cline{1-11}
\hline
\end{tabular}
\label{tab:open_ter}
\end{table*}

\subsection{Environment Setup}
Our experimental study consists of two main parts. 
% In the first section, we delve into quantitative research on the interference of robotic navigation with individual pedestrians across various scenarios, focusing on safety measures taken by the robot in response to sudden pedestrian behavior. 
The first section is dedicated to exploring the interplay between robots and human traffic in geometrically constrained scenarios, including bottleneck areas, foot traffic intersections, and narrow corridors. The second part broadens our scope to interactions between robots and dense crowds in open terrain environments. The simulated environments relied on the PedSim simulation platform.
% In conducting the first part, we utilized the ORCA interaction simulator, while the latter two parts relied on the PedSim simulation platform.

\subsection{Metric Setup}
For the interaction between robots and pedestrian flows, we have proposed the following nine metrics:
\begin{itemize}
\item Complete Ratio: Percentage of total path completed by the robot.
\item Success Rate: Percentage of successful robot arrivals at the destination out of the total number of experiments.
\item Timeout Rate: Percentage of robot timeouts out of the total number of experiments.
\item Collision Rate: Percentage of instances where the distance between the robot and pedestrians is less than the safety distance, combined with the number of instances where the robot collides with walls, out of the total number of experiments.
\item Freezing Number: Number of instances where the robot has to stop for more than 3 seconds due to dense pedestrian flow.
\item Jerk: The cumulative average of jerk per episode during robot navigation.
\item Frontal Interaction Times \cite{predhumeau2021agent}: Count how many times there are dangerous interactions between the robot and pedestrians based on this metric. It expressed the number of face-to-face interactions between the robot and people within a range of 1.5 meters.
\item Cumulative Density : The cumulative value of pedestrian density within 0.5 meters directly in front of the robot for each episode. 
\item Execution Time: The time duration required for the robot to travel from the starting point to the destination in each episode.
\end{itemize} 

% In order to investigate the performance of robots and individual pedestrians at a micro-level, we have additionally introduced two metrics:
% \begin{itemize}
% \item Human Path Deviation: The positional deviation of individual pedestrians at each moment, considering both the presence and absence of the robot.
% \item Success Rate of Passive Safety: Ability to safely perform emergency stops in response to other's abnormal behavior. 
% \end{itemize}

\subsection{Baseline Setup}
We use the following six algorithms as baselines.
\begin{itemize}
\item Optimal Reciprocal Collision Avoidance (ORCA) \cite{van2010optimal}: A distributed algorithm suitable for large-scale multi-agent navigation, it effectively avoids collisions and can quickly compute in real-time environments.
\item Distribution Navigation (DistNav) \cite{sun2021move}: An algorithm that utilizes multi-agent game theory for optimization and cooperative navigation.
\item Dynamic Window Approach (DWA) \cite{4209377}: The algorithm defines a "dynamic window" in the motion space, takes into account the dynamic constraints and environmental conditions for velocity sampling, and uses an evaluation function to select the optimal velocity and steering commands.
\item Risk RRT \cite{fulgenzi2010risk}: An algorithm based on the basic principles of the RRT algorithm and introduces risk assessment during the path search process. It uses random sampling and tree growth to search for feasible paths.
\item Dynamic Channel \cite{8794192}: This algorithm combines topological path planning with low-level motion planning to address the challenges of real-time navigation in dense human environments.
\item Crowd Planner \cite{fan2021crowddriven}: This algorithm considers the velocity field of pedestrians and utilizes it to adjust the cost during path search.
\end{itemize}

% \subsection{Interactions of individual pedestrians}
% We commence by investigating the influence of robots on individual pedestrians. Specifically, we quantify the deviations in each pedestrian's displacement when a robot is present versus when it is absent. Using the ORCA simulation tool, two experimental conditions were designed: one with the robot navigating normally and another with the robot removed, ensuring pedestrians start from identical points and aim for the same destination. The deviation of each pedestrian's path is calculated under both conditions, with the assumption that other pedestrians maintain consistent strategies.
% Drawing from [1], we adopt the principle of 'passive safety' to evaluate how effectively a robot handles situations involving unexpected pedestrian actions. This principle dictates that, should a pedestrian suddenly veer towards the robot, the robot is programmed to safely halt to avert potential collisions, even if evasion isn't feasible in a timely manner. It thereby ensures that, in the event of a collision, the cause is attributed to the failure of other agents to stop, rather than the robot's malfunction.

\subsection{Interactions with Pedestrian Flow in geometrically constrained scenarios}

Dense pedestrian movements often occur due to geometrically constrained scenarios, where constraints of physical space inevitably lead to increased pedestrian density. Our primary focus lies in understanding the interaction between robots and pedestrian flows within three specific geometrically restricted terrains:

\begin{itemize}
\item \textbf{Narrow Corridors (NC)}:
Here, robots must navigate pedestrian flows within confined pathways. The restricted space poses a limitation on the maneuverability of both parties to avoid collisions.
\item \textbf{Foot-traffic Intersections (FI)}:
At these junctures, multiple pedestrian flows intersect, and robots must manage not only the forward and backward flows but also consider cross-traffic. The multi-directional nature of these pedestrian flows intensifies the complexity of robot navigation.
\item \textbf{Bottleneck Areas (BA)}:
Pedestrian flows converge from wider spaces into constrained zones, such as entrances and exits. These areas often witness traffic congestion, and it's crucial for robots to traverse these points without impeding the flow of traffic.
\end{itemize}

Concerning the interaction patterns between robots and pedestrian flows, we mainly explore the following modes:

\begin{itemize}
\item \textbf{Downstream Travel (DT) - No Minor Counterflow}:
This mode depicts the robot moving synchronously with the main flow, with no opposing minor flow.
\item \textbf{Downstream Travel (DT)  - With Minor Counterflow}:
This mode depicts the robot moving along with the main flow but occasionally confronting minor opposing flows.
\item \textbf{Upstream Travel (UT)  - No Minor Counterflow}:
This mode depicts the robot moving against the main flow, without the presence of opposing minor flows.
\item \textbf{Upstream Travel (UT)  - With Minor Counterflow}:
This mode depicts the robot moving against the main flow, occasionally meeting minor opposing flows.
\end{itemize}

We systematically combined these geometric scenarios with the interaction modes, resulting in a total of 12 experimental configurations. Each configuration was repeated 40 times to ensure the robustness and reliability of our findings.
The results are shown in Table \ref{tab:no_counter} and \ref{tab:with_counter}

\subsection{Interactions with Dense Crowds in Open Terrain}
In Interactions with Dense Crowds in Open Terrain, dense crowds primarily result from a large number of individuals rather than geometric constraints. While robots are no longer strictly limited by static obstacles, this also means that pedestrian flows are no longer as predictable as before. For instance, the flow of people at the same location may vary in different directions at different times, which adds complexity to the robot's long-distance navigation.
The results are shown in Table \ref{tab:open_ter}.

\subsection{Ablation Study}
\begin{figure}[h]
    \centering
    \includegraphics[width=\columnwidth]{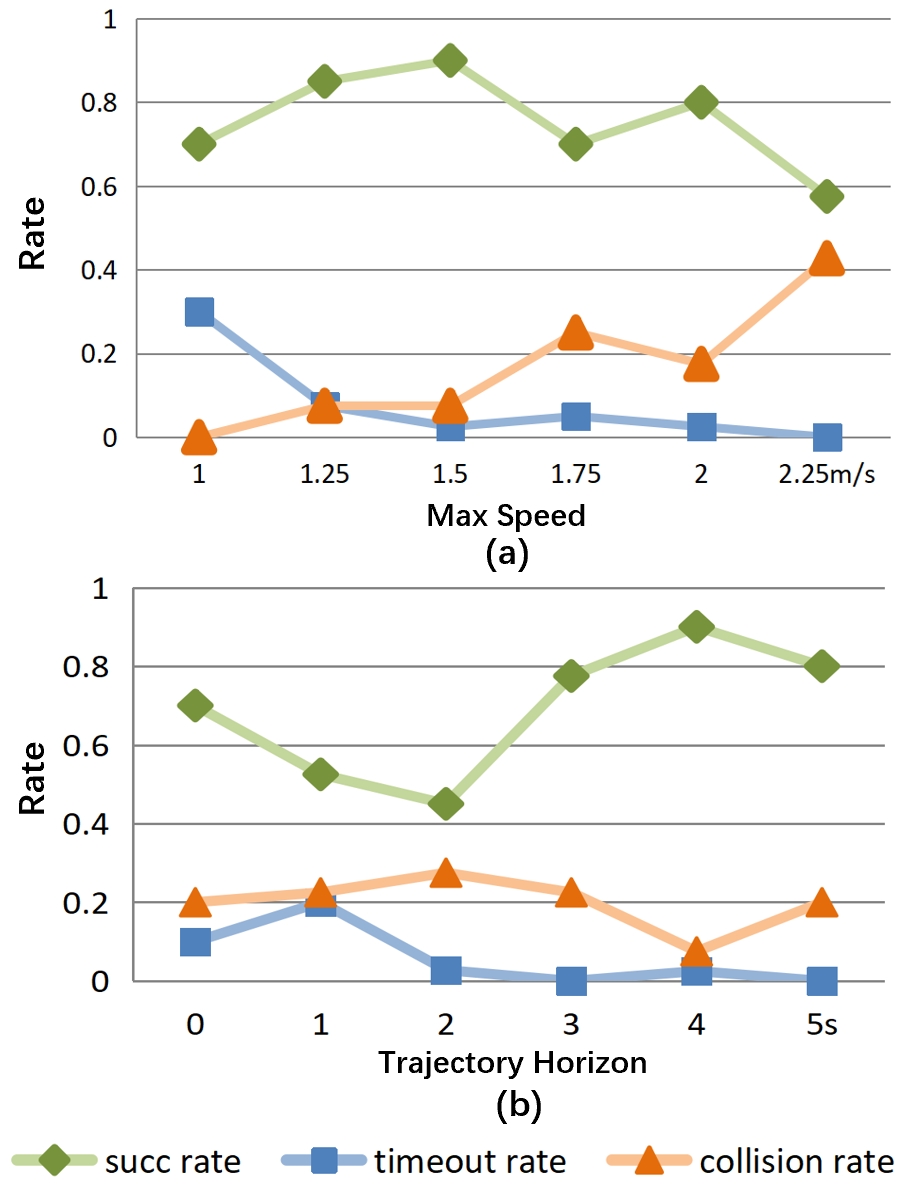}
    \caption{Ablation study is conducted to examine the variations in success rate, timeout rate, and collision rate of trajectories under different maximum speeds and durations. We evaluate the performance of the trajectories by testing them in various scenarios with different parameter settings.}
    \label{fig:abl_zhexian}
\end{figure}
\begin{figure}[h]
    \centering
    \includegraphics[width=\columnwidth]{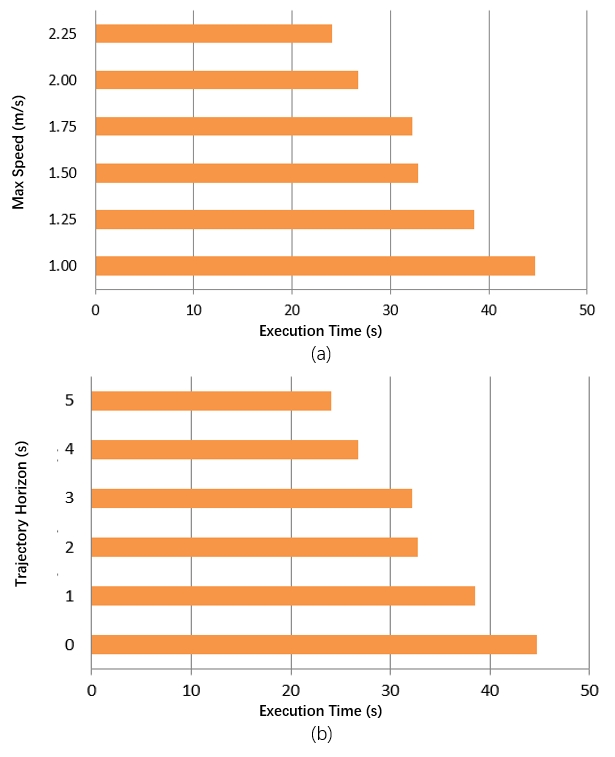}
    \caption{The ablation study aims to analyze the influence of different maximum speeds and durations on the average time taken for the robot to successfully reach the endpoint of trajectories. }
    \label{fig:abl_single_bar}
\end{figure}
\begin{figure}[h]
    \centering
    \includegraphics[width=\columnwidth]{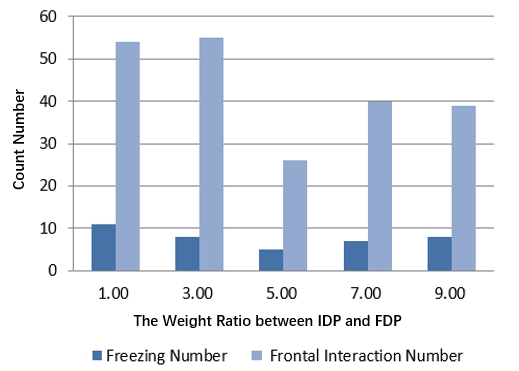}
    \caption{The figure illustrates the relationship between different proportions of individual disturbance and flow disturbance and their impact on the occurrences of "freezing" and "frontal interactions" by the robot. }
    \label{fig:abl_double_bar}
\end{figure}
In our ablation study, we systematically explore the impact of varying the parameters of partial motion trajectories, specifically, maximum speed and trajectory horizon, on the performance of our algorithm. By manipulating these variables, we aim to delineate the sensitivity of our algorithm's efficacy with respect to changes in dynamic motion constraints and predictive time frames. 
Figure \ref{fig:abl_zhexian} and \ref{fig:abl_single_bar} present the main findings of the study. Figure \ref{fig:abl_zhexian} examines the success rate, collision rate, and timeout rate, providing insights into system performance. Figure \ref{fig:abl_single_bar} focuses on the time taken to reach the endpoint successfully.

The study initially investigates the effects of varying maximum speeds, ranging from 1m/s to 2.25m/s, on partial motion trajectories, as illustrated in \ref{fig:abl_zhexian}(a) and Figure \ref{fig:abl_single_bar}(a). The success rate exhibits a gradual increase within the 1-1.5m/s range, reaching an optimal performance level. However, a subsequent decline is observed at 1.75m/s, with a transient increase noted at 2m/s before displaying an overall decreasing trend. The collision rate demonstrates an ascending pattern, while the timeout rate shows a descending tendency. The execution time metric demonstrates a positive correlation between speed and shorter completion times. However, this relationship is not linear due to the dynamic interactions with the crowd, which may impede the robot's continuous advancement at its maximum speed.

Subsequently, we delve into the examination of the duration of the partial motion trajectory, encompassing a range of 0 to 5 seconds. It is noteworthy that the special case of 0 seconds entails the absence of explicit modeling of individual disturbance, wherein the modeling solely encompasses the overall impact of flow disturbance. In contrast, the subsequent duration of 1 to 5 seconds entails a comprehensive investigation into the incorporation of duration when modeling individual disturbance within the trajectory.
The results, as depicted in Figure \ref{fig:abl_zhexian}(b) and Figure \ref{fig:abl_single_bar}(b), indicate that even without considering Individual Disturbance, the algorithm achieves an approximate success rate of $70\%$. However, excessively short durations of Individual Disturbance tend to diminish the algorithm's performance, as close-range interactions pose potential risks. Notably, the success rate significantly improves beyond 3 seconds and reaches its peak at 4 seconds. On the other hand, trajectories with excessively long durations, such as exceeding 5 seconds, may introduce challenges in terms of increased spatial length, thereby complicating the representation of trajectory diversity and optimality.
In terms of execution duration, trajectories with longer horizons tend to have shorter overall time consumption.

Furthermore, we examined the impact of the proportion between Individual Disturbance and Flow Disturbance on performance. Our primary focus centered on two key performance indicators: the Freezing Number, representing the count of instances when the robot becomes immobilized for a defined duration within the crowd, and the number of Frontal Interactions, which quantifies close-range disruptions caused by the robot's interaction with individuals in the crowd. A higher frequency of Frontal Interactions signifies a greater influence on the affected individuals.
We established a range of proportions from 1 to 9 and conducted a series of 40 experimental trials. The outcomes of these experiments are presented in Figure \ref{fig:abl_double_bar}. The analysis reveals a consistent pattern in the behavior of both indicators, displaying an initial decline followed by an upward trend. The proportion of 5.0 exhibits the lowest frequency, indicating the highest level of performance efficacy.

\section{Real World Experiment Evaluation}
\label{sec:real_world_experiment}
\begin{figure}[t]
    \centering
    \includegraphics[width=\columnwidth]{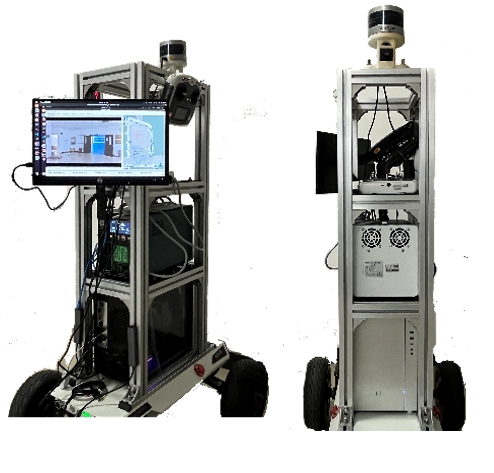}
    \caption{The picture of Hunter-se robot with Ackerman motion structure. We performed calibration on the radar and camera and mounted them at a height of approximately 1.4m. The Inertial Measurement Unit is fixed at the center of the robot and used for aiding in localization.}
    \label{fig:robot_figure}
\end{figure}
% \begin{figure}[t]
%     \centering
%     \includegraphics[width=\columnwidth]{picture/figure6_experiment_scenario.png}
%     \caption{The experimental scenario consists of an asphalt road with a length of approximately 119 meters and a width of around 3 meters. This road permits the passage of pedestrians and electric bicycles, while excluding cars.}
%     \label{fig:experiment_environment}
% \end{figure}

\begin{figure}[t]
    \centering
    \includegraphics[width=\columnwidth]{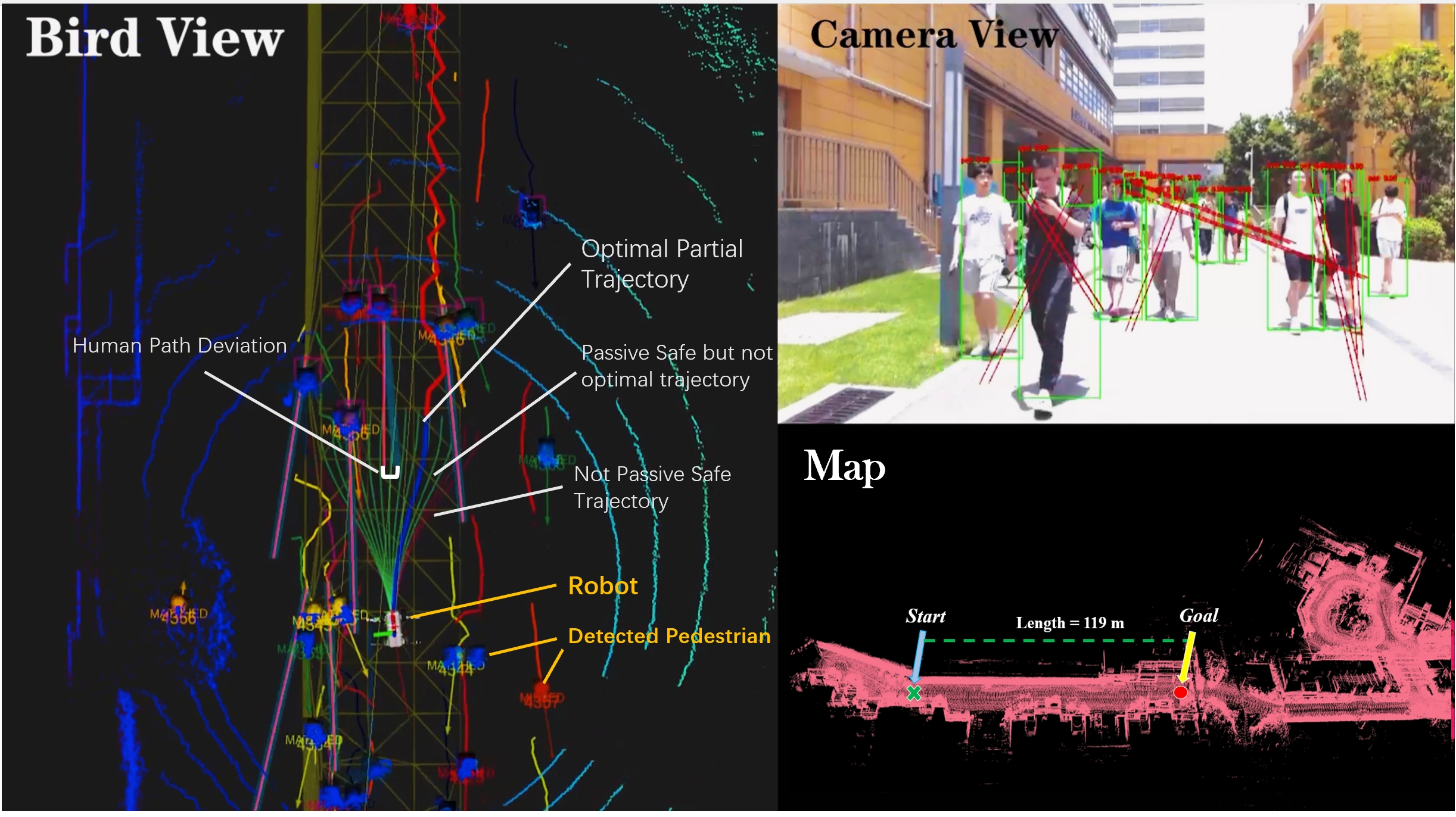}
    \caption{The experimental scenario consists of an asphalt road with a length of approximately 119 meters and a width of around 3 meters. This road permits the passage of pedestrians and electric bicycles, while excluding cars.}
    \label{fig:experiment_environment}
\end{figure}

\subsection{Robot Setup}
We utilized a Hunter-se robot, incorporating an Ackerman motion structure, as our experimental platform. To ensure accurate measurements, we performed calibration on both the radar and camera systems, positioning them at a height of approximately 1.4m. The chosen height serves two purposes: firstly, it enhances the visibility of the robot to pedestrians, particularly in dense crowds; secondly, it facilitates the extraction of gaze information from nearby pedestrians, which is crucial for modeling disturbances in the context of disturbance modeling.
The Inertial Measurement Unit (IMU) was precisely mounted at the robot's center, providing valuable data for localization tasks. Figure \ref{fig:robot_figure} is a visual representation of the robot setup.

\begin{table*}[tbhp]
\centering
\renewcommand{\arraystretch}{1.5}
\caption{Real world crowd analysis and our  algorithm performance.}
\label{tab: real_world_data}
\begin{tabular}{
    p{0.10\textwidth}  % Scenario column
    p{0.10\textwidth}  % Pattern column
    p{0.15\textwidth}  % Total column
    p{0.07\textwidth}  % Success Rate column
    p{0.07\textwidth}  % Timeout Rate column
    p{0.07\textwidth}  % Collision Rate column
    p{0.06\textwidth}  % Freezing Num column
    p{0.06\textwidth}  % Frontal Interact Num column
    p{0.09\textwidth}  % Execute Time column
}
\hline
\multirow{2}{*}{Scenario} & \multirow{2}{*}{Total Ped Count} & \multirow{2}{*}{Minor Flow Ped Count} & \multicolumn{6}{c}{Metrics} \\
\cline{4-9}
& & & Success Rate (\%) & Timeout Rate (\%) & Collision Rate (\%) & Freezing Num & Frontal Interact Num & Avg Execute Time (s) \\
\hline
Downstream & 463 & 114 & 100.00 & 0.00 & 0.00 & 0 & 0 & 147.500 \\
Upstream  & 1061   & 51 & 83.33 & 16.67 & 0.00 & 5 & 36 &151.533 \\
\cline{1-9}
\hline
\end{tabular}
\label{tab:my_table}
\end{table*}

\subsection{Perception Algorithms Setup}
\subsubsection{PointCloud clustering}
We initiate the process by employing ground removal techniques to eliminate the points corresponding to the ground and walls from the sensor's point cloud data. This step serves the purpose of reducing computational overhead and performing an initial candidate selection. Subsequently, the remaining point cloud data is utilized for feature extraction and classification. Drawing inspiration from the methodology proposed in \cite{6943142}, we adopt the dp-means algorithm for cluster analysis. The clustering process commences with a single cluster and incrementally increases the cluster number when encountering excessively large clusters. Furthermore, we incorporate a centroid-based distance evaluation to facilitate the merging of smaller clusters, aiming to achieve precise separation of pedestrians within dense crowd environments.

In this context, we establish two distinct types of clustering: firstly, a fine-grained point cloud clustering approach, aimed at seamless integration with image data; and secondly, a stringent clustering methodology that incorporates rigorous criteria for scale and distance, specifically tailored for independent 3D point cloud detection.

\subsubsection{Image-based Pedestrian Detection}
We employed the YOLOv5 image detection algorithm, specifically utilizing the YOLOv5n model, in conjunction with TensorRT for accelerated neural network processing. Leveraging visual imagery, we obtained precise 2D coordinates corresponding to each pedestrian within the image. Subsequently, we projected the clustering centroids, derived from the previous section, onto the image plane utilizing the extrinsic matrix encompassing the radar and camera parameters. By establishing a one-to-one correspondence with the 2D pedestrian coordinates, we successfully derived the accurate 3D coordinates of the pedestrians.

\subsubsection{Radar-based Pedestrian Detection}
Due to the limited field of view of the camera, visual-based methods alone are insufficient for comprehensive pedestrian detection across all directions. Therefore, we integrated a complementary pure point cloud-based detection approach. Leveraging a support vector machine, we conducted classification of point cloud features, including slice features that characterize the contour of pedestrians based on height specification, and reflectance intensity distribution features to discern pedestrian-related point clouds. By inputting the processed clusters from the previous section into the classifier, we obtained detections of 3D coordinates from diverse directions.

\subsubsection{Pedestrian Detection Aggregating and Tracking}

In order to synergistically integrate detection information derived from multiple methods, we adopted a fusion strategy, as outlined in reference \cite{6916032}, wherein the output results of the individual detectors were subjected to a combination process prior to being fed into the tracking module. In instances where there were overlapping regions identified by both detectors, we employed the technique of non-maximum suppression to effectively merge the detection outcomes and yield a consolidated result.

\subsubsection{Face Detection and Gaze Estimation}
To facilitate our gaze detection process, we employed the YOLO-Face algorithm for accurate face detection. Subsequently, the detected facial pixels were fed into the openvino library, enabling the implementation of gaze detection techniques. Specifically, we utilized the head-pose-estimation-adas-0001 model to estimate the precise pose information of the head, providing valuable insights into its orientation. Additionally, the facial-landmarks-35-adas-0002 model was employed to precisely identify and locate facial landmarks, contributing to the accurate estimation of gaze direction. Finally, the open-closed-eye-0001 model was utilized to determine the visual focus by assessing the status of the eyes. This comprehensive approach ensured reliable and accurate gaze detection within our system.

\begin{figure*}[h]
    \centering
    \includegraphics[width=\textwidth]{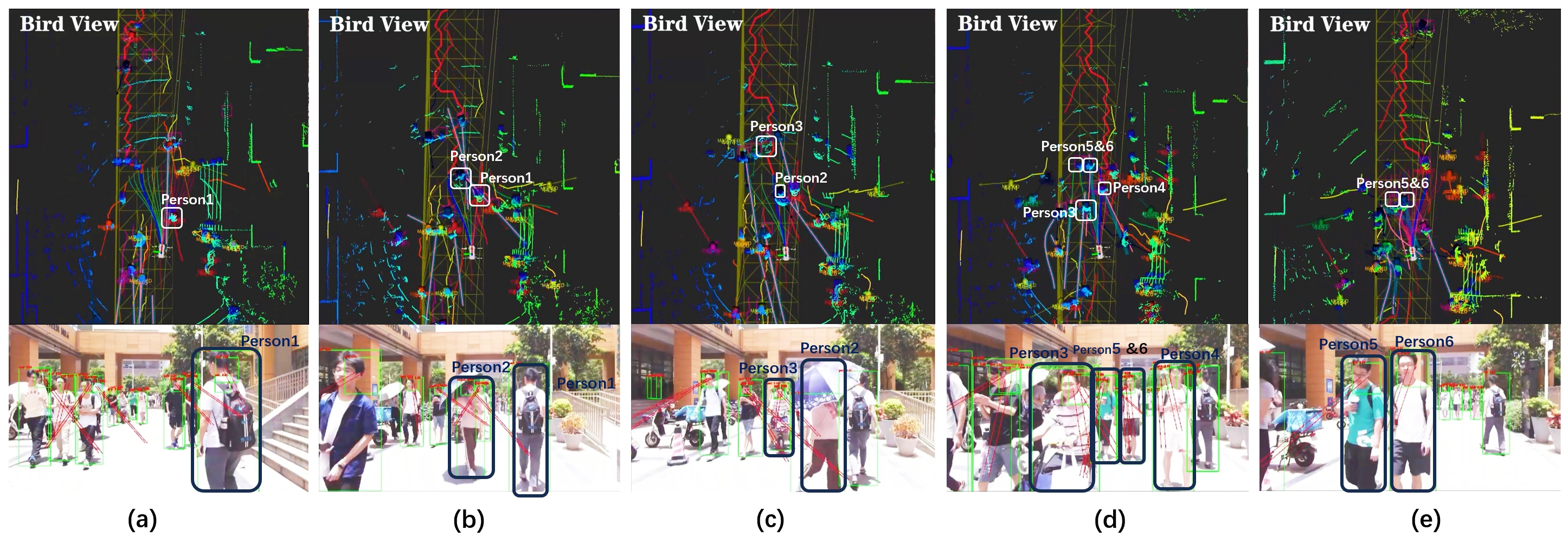}
    \caption{Systematic Management of Recurrent Emergencies. (a) The robot encounters an obstruction by pedestrian 1 ahead and makes a deliberate decision not to interfere with their trajectory, opting instead to circumnavigate around them. (b) Pedestrian 2 abruptly appears from a concealed area and approaches the robot. The robot promptly opts to bypass pedestrian 2 from the left side. (c) With their coordinated efforts, the interaction between the robot and pedestrian 2 is reaching its conclusion, while pedestrian 3 on an electric bicycle is approaching. (d) The robot decides to pass through the space between pedestrian 3 and pedestrian 4. Pedestrians 5 and 6, on the other hand, are heading towards the robot from a distance. (e) The robot successfully navigates through the congested area and now has ample space. At this point, it decides to bypass pedestrians 5 and 6 directly, avoiding any further interaction with them.}
    \label{fig:real_world_emergencies_demo}
\end{figure*}
\begin{figure*}[h]
    \centering
    \includegraphics[width=\textwidth]{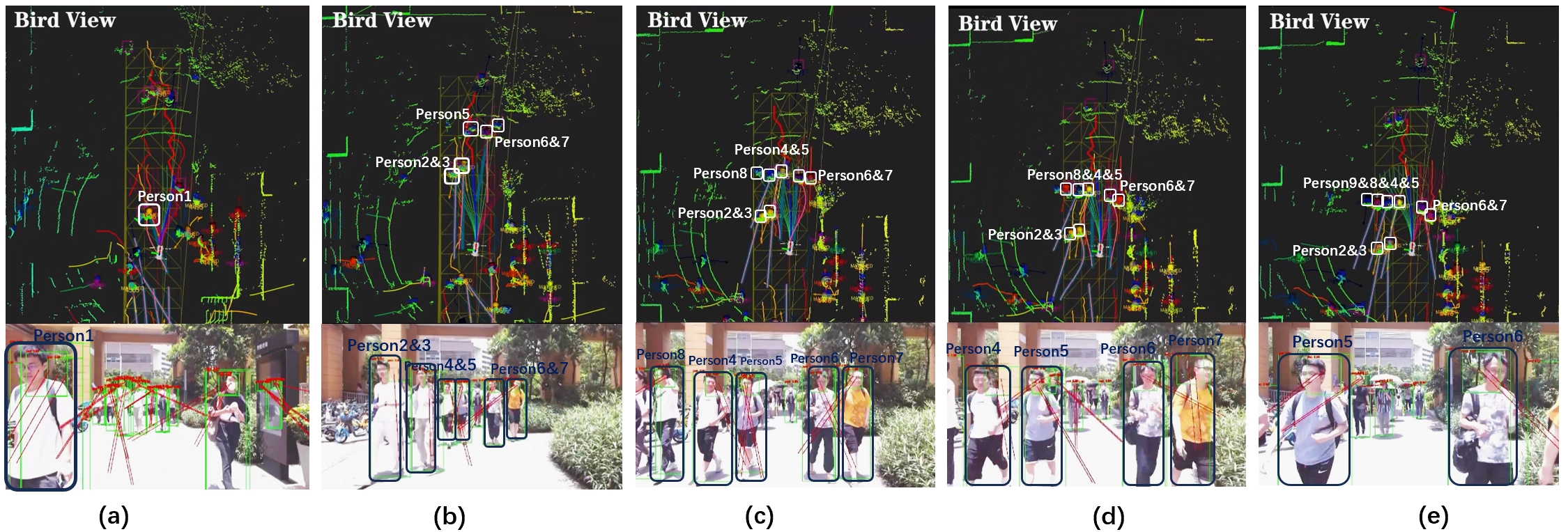}
    \caption{Negotiation with Pedestrians in Constrained Space Environments. (a) When ample space is available, the robot opts for a direct circumvention of pedestrian 1. (b) The robot strategically maneuvers by positioning itself behind pedestrians 2 and 3 and executing a left-side circumvention, as well as bypassing pedestrians 5, 6, and 7. (c) The robot identifies other individuals (person8) to the left of pedestrians 4 and 5, rendering a direct bypass infeasible due to insufficient space. Consequently, it opts to navigate through the narrow gap between pedestrians 5 and 6. (d-e) The robot maintains a persistent and coherent expression of its intentions, eliciting a collaborative response from pedestrians 5 and 6. Consequently, the interstitial space between pedestrians 5 and 6 progressively expands.}
    \label{fig:real_world_negotiate_demo}
\end{figure*}

\subsection{Minimal Intrusive Planning}
We employed the Minimal Intrusive Planning algorithm with a maximum velocity constraint of 1.2m/s and a planning horizon of 4 seconds. We employed the Model Predictive Control (MPC) approach to track trajectories at a frequency of approximately 10Hz. The MPC algorithm utilized a planning horizon of 20 steps, with each step discretized by 0.1s.
We selected a pedestrian pathway between two canteens as the experimental setting, with an approximate length of 119 meters and a width of approximately 3 meters. This pathway allows for the passage of pedestrians and electric bicycles, while excluding automobiles.
Figure \ref{fig:experiment_environment} illustrates the scene of the experiment.
For both pedestrians and cyclists, we treated them equally as pedestrians, with the only distinction being the preferred operating speed.

We conducted 12 experiments in total, divided equally into six groups for downstream and upstream directions. The interaction data is presented in Table \ref{tab: real_world_data}, showing the total number of pedestrians involved in the interactions (second column) and the count of pedestrians moving against the flow during those interactions (third column).

In the downstream scenario, our algorithm exhibited a $100\%$ success rate, with zero occurrences of freezing and frontal interactions, showcasing the harmonious interaction between the robot and the crowd. In the upstream scenario, the algorithm achieved an 83.$33\%$ success rate and a $16.67\%$ timeout rate. Notably, the number of freezing instances and frontal interactions remained low at 5 and 36, respectively. Although the execution time was slightly longer than in the downstream scenario due to the occurrence of timeouts, it is worth mentioning that our algorithm recorded no instances of collisions. This underscores the safety aspect of the minimally intrusive algorithm in navigating through dense crowds.

\subsection{Demonstration}

To substantiate the empirical viability of our advanced system, designed for unobtrusive navigational guidance within authentic environmental contexts, we elucidate the system's applicability through two distinct demonstrations: the adept orchestration of navigation during frequent emergency situations and the diplomatic spatial negotiation with pedestrians. These are depicted in Figures \ref{fig:real_world_emergencies_demo} and \ref{fig:real_world_negotiate_demo}, respectively.

The visual exposition commences with a bird-eye view representation, delineating both the pedestrian coordinates in a three-dimensional construct and the autonomous agent's locale, as illustrated through a point cloud rendition in the upper echelon of the visual array. Complementing this perspective, the subsequent echelon provides a series of images captured from the robot's egocentric viewpoint for corroborative elucidation.

The narrative of the first scenario, involving recurrent emergency management, unfolds in Figure \ref{fig:real_world_emergencies_demo}. Herein, the robot encounters a series of pedestrian interactions, each managed with meticulous strategic forethought. Initially, upon encountering pedestrian 1, as depicted in Figure \ref{fig:real_world_emergencies_demo}(a), the robot exhibits restraint in preserving the pedestrian's undisturbed path, opting instead for a navigational detour. A spontaneous appearance by pedestrian 2 from a blind spot, as shown in Figure \ref{fig:real_world_emergencies_demo}(b), necessitates an agile lateral evasion by the robot. As this interaction reaches a collaborative denouement, as depicted in Figure \ref{fig:real_world_emergencies_demo}(c-d), the emergence of pedestrian 3 on an electric bicycle is noted. Capitalizing on the spatial opening, the robot adeptly maneuvers between pedestrians 3 and 4. Subsequently, as pedestrians 5 and 6 begin to converge on the robot's trajectory from afar, the robot, now immersed in a less constricted space as showcased in Figure \ref{fig:real_world_emergencies_demo}(e), judiciously elects a direct overtaking course, thus circumventing any further pedestrian engagements.

In scenarios characterized by spatial restrictions and pedestrian negotiations, the robot demonstrates a refined adaptability in its traversal approach. Presented initially with a generous spatial allowance, as captured in Figure \ref{fig:real_world_negotiate_demo}(a), the robot proficiently selects an immediate bypassing route to circumvent pedestrian 1. Progressing to Figure \ref{fig:real_world_negotiate_demo}(b), the robot employs sagacious positioning tactics, aligning itself behind pedestrians 2 and 3 and executing an astute leftward circumnavigation to overtake pedestrians 5, 6, and 7. Advancing through the scenario, the robot prudently discerns the proximity of an additional individual, person 8, as visualized in Figure \ref{fig:real_world_negotiate_demo}(c), and, recognizing the spatial constraints, opts for a precise traversal through the constricted aperture between pedestrians 5 and 6. Throughout these engagements, the robot consistently upholds a clear and deliberate conveyance of its navigational intent, thereby successfully invoking a cooperative demeanor among the pedestrians. This dynamic facilitates a progressive dilation of the gap between pedestrians 5 and 6, culminating in the robot's unimpeded passage as encapsulated in Figure \ref{fig:real_world_negotiate_demo}(c-e).

\section{Conclusions}
In this paper, we propose two penalty terms for measuring the disturbance of vehicle trajectory to pedestrains.
To generate a vehicle trajectory with these terms in real time, we build a abstract representation of the environment with flow map and use sampling-based search method to find best trajectory.
In experiment section, we proves that our method outperform baseline methods in both geometrically constrained scenarios and open terrains.
For future directions, we may focus on integrating human-in-the-loop reinforcement learning approaches with advanced large-scale language models to fortify the theoretical framework of minimal intrusive navigation.

\bibliographystyle{IEEEtran}
\bibliography{ref.bib}

% \newpage





\vfill

\end{document}